\pdfoutput=1

\documentclass[10pt,twocolumn,letterpaper]{article}
\usepackage{cvpr}
\makeatletter
\@namedef{ver@everyshi.sty}{}
\makeatother
\usepackage{times}
\usepackage{epsfig}
\usepackage{graphicx}
\usepackage{amsmath}
\usepackage{amssymb}
\usepackage{cite}

\usepackage[toc,page]{appendix}

\usepackage{caption,array,tikz}

\newcommand{\R}{{\mathbb{R}}}

\newcommand\mypara[1]{\vspace{0.2cm}\noindent \textbf{#1} \hspace{0.2cm}}

\usepackage[breaklinks=true,bookmarks=false]{hyperref}

\cvprfinalcopy 

\def\cvprPaperID{****} 

\ifcvprfinal\fi
\title{A Morphable Face Albedo Model}

\author{William A.~P.~Smith$^1\quad$ Alassane Seck$^{2,3}\quad$ Hannah Dee$^3$ \\Bernard Tiddeman$^3\quad$ Joshua Tenenbaum$^4\quad$ Bernhard Egger$^4$\\
{\small $^1$University of York, UK$\quad$ $^2$ARM Ltd, UK$\quad$ $^3$Aberystwyth University, UK$\quad$ $^4$MIT - BCS, CSAIL \& CBMM, USA}\\
{\tt\small william.smith@york.ac.uk}, {\tt\small alou.kces@live.co.uk}, {\tt\small \{hmd1,bpt\}@aber.ac.uk}, {\tt\small \{jbt,egger\}@mit.edu}
}

\begin{document}

\twocolumn[{%
\renewcommand\twocolumn[1][]{#1}%
\maketitle \thispagestyle{empty} \vspace{-0.7cm}
\begin{center}
    \centering
\includegraphics[trim=0px 168px 0px 0px,width=16cm]{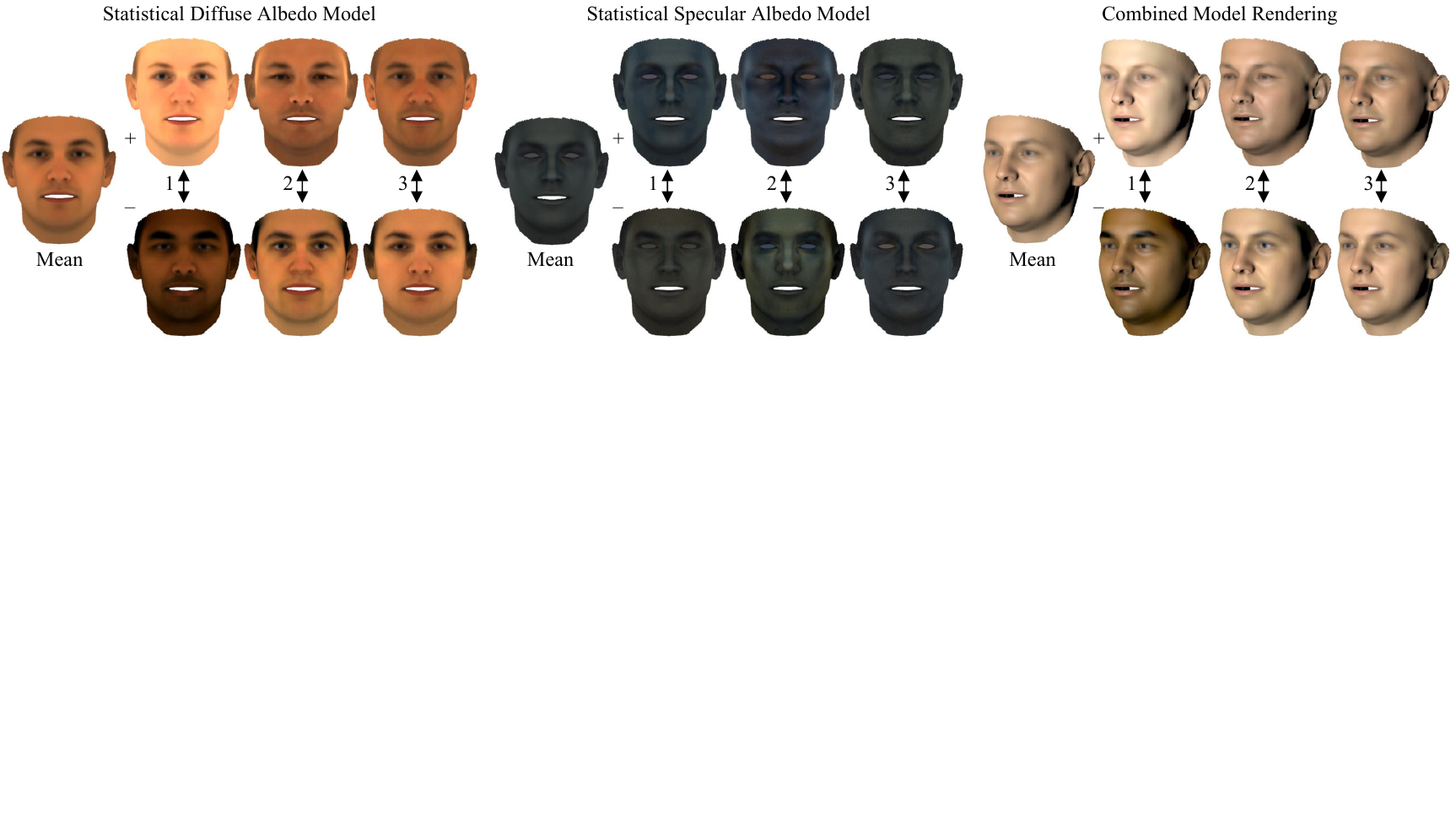}\vspace{-0.3cm}
    \captionof{figure}{First 3 principal components of our statistical diffuse (left) and specular (middle) albedo models. Both are visualised in linear sRGB space. Right: rendering of the combined model under frontal illumination in nonlinear sRGB space.}
    \label{fig:teaser}
\end{center}%
}]

\begin{abstract}
In this paper, we bring together two divergent strands of research: photometric face capture and statistical 3D face appearance modelling. We propose a novel lightstage capture and processing pipeline for acquiring ear-to-ear, truly intrinsic diffuse and specular albedo maps that fully factor out the effects of illumination, camera and geometry. Using this pipeline, we capture a dataset of 50 scans and combine them with the only existing publicly available albedo dataset (3DRFE) of 23 scans. This allows us to build the first morphable face albedo model. We believe this is the first statistical analysis of the variability of facial specular albedo maps. This model can be used as a plug in replacement for the texture model of the Basel Face Model (BFM) or FLAME and we make the model publicly available. We ensure careful spectral calibration such that our model is built in a linear sRGB space, suitable for inverse rendering of images taken by typical cameras. We demonstrate our model in a state of the art analysis-by-synthesis 3DMM fitting pipeline, are the first to integrate specular map estimation and outperform the BFM in albedo reconstruction. 
\end{abstract}


\vspace{-0.5cm}
\section{Introduction}

3D Morphable Models (3DMMs) were proposed over 20 years ago \cite{BlanzVetter1999} as a dense statistical model of 3D face geometry and texture. They can be used as a generative model of 2D face appearance by combining shape and texture parameters with illumination and camera parameters that are provided as input to a graphics renderer. Using such a model in an analysis-by-synthesis framework allows a principled disentangling of the contributing factors of face appearance in an image. More recently, 3DMMs and differentiable renderers have been used as model-based decoders to train convolutional neural networks (CNNs) to regress 3DMM parameters directly from a single image \cite{tewari17MoFA}.

The ability of these methods to disentangle intrinsic (geometry and reflectance) from extrinsic (illumination and camera) parameters relies upon the 3DMM capturing only intrinsic parameters, with geometry and reflectance modelled independently. 3DMMs are usually built from captured data \cite{BlanzVetter1999,paysan20093d,LSFM2016,LYHM2017}. This necessitates a face capture setup in which not only 3D geometry but also intrinsic face reflectance properties, e.g.~diffuse albedo, can be measured. A recent large scale survey of 3DMMs \cite{egger20193d} identified a lack of intrinsic face appearance datasets as a critical limiting factor in advancing the state-of-the-art. Existing 3DMMs are built using ill-defined ``textures'' that bake in shading, shadowing, specularities, light source colour, camera spectral sensitivity and colour transformations. Capturing truly intrinsic face appearance parameters is a well studied problem in graphics but this work has been done largely independently of the computer vision and 3DMM communities.

In this paper we present a novel capture setup and processing pipeline for measuring ear-to-ear diffuse and specular albedo maps. We use a lightstage to capture multiple photometric views of a face. We compute geometry using uncalibrated multiview stereo, warp a template to the raw scanned meshes and then stitch seamless per-vertex diffuse and specular albedo maps. We capture our own dataset of 50 faces, combine this with the 3DRFE dataset~\cite{stratou2011effect} and build a statistical albedo model that can be used as a drop-in replacement for existing texture models. We make this model publicly available. To demonstrate the benefits of our model, we use it with a state-of-the-art fitting algorithm and show improvements over existing texture models.

\subsection{Related work}

\mypara{3D Morphable Face Models} The original 3DMM of Blanz and Vetter \cite{BlanzVetter1999} was built using 200 scans captured in a Cyberware laser scanner which also provides a colour texture map. Ten years later the first publicly available 3DMM, the Basel Face Model (BFM) \cite{paysan20093d}, was released. Again, this was built from 200 scans, this time captured using a structured light system from ABW-3D. Here, texture is captured by three cameras synchronised with three flashes with diffusers, providing relatively consistent illumination. The later BFM 2017 \cite{gerig2018morphable} used largely the same data from the same scanning setup. More recently, attempts have been made to scale up training data to better capture variability across the population. Both the large scale face model (LSFM) \cite{LSFM2016} (10k subjects) and Liverpool-York Head Model (LYHM) \cite{LYHM2017} (1.2k subjects) use shape and textures captured by a 3DMD multiview structured light scanner under relatively uncontrolled illumination conditions. Ploumpis~\etal~\cite{ploumpis2019combining} show how to combine the LSFM and LYHM but do so only for shape, not for texture. All of these previous models use texture maps that are corrupted by shading effects related to geometry and the illumination environment, mix specular and diffuse reflectance and are specific to the camera with which they were captured. Gecer~\etal~\cite{gecer2019ganfit} use a Generative Adversarial Network (GAN) to learn a nonlinear texture model from high resolution scanned textures. Although this enables them to capture high frequency details usually lost by linear models, it does not resolve the issues with the source textures.

Recently, there have been attempts to learn  3DMMs directly from in-the-wild data simultaneously with learning to fit the model to images \cite{Tran2018,t19fml}. The advantage of such approaches is that they can exploit the vast resource of available 2D face images. However, the separation of illumination and albedo is ambiguous while non-Lambertian effects are usually neglected and so these methods do not currently provide intrinsic appearance models of a quality comparable with those built from captured textures.

\mypara{Face Capture} Existing methods for face capture fall broadly into two categories: photometric and geometric. Geometric methods rely on finding correspondences between features in multiview images enabling the triangulation of 3D position. These methods are relatively robust, can operate in uncontrolled illumination conditions, provide instantaneous capture and can provide high quality shape estimates \cite{Beeler:10}. They are sufficiently mature that commercial systems are widely available, for example using structured light stereo, multiview stereo or laser scanning. However, the texture maps captured by these systems are nothing other than an image of the face under a particular set of environmental conditions and hence are useless for relighting. Worse, since appearance is view-dependent (the position of specularities changes with viewing direction), no one single appearance can explain the set of multiview images.

On the other hand, photometric analysis allows estimation of additional reflectance properties such as diffuse and specular albedo \cite{ma2007rapid}, surface roughness \cite{Ghosh:09} and index of refraction \cite{Ghosh:10} through analysis of the intensity and polarisation state of reflected light. This separation of appearance into geometry and reflectance is essential for the construction of 3DMMs that truly distentangle the different factors of appearance. The required setups are usually much more restrictive, complex and not yet widely commercially available. Hence, the availability of datasets has been extremely limited, particularly of the scale required for learning 3DMMs. There is a single publicly available dataset of scans, the 3D Relightable Facial Expression (3DRFE) database \cite{stratou2011effect} captured using the setup of Ma~\etal~\cite{ma2007rapid}.

Ma~\etal~\cite{ma2007rapid} were the first to propose the use of polarised spherical gradient illumination in a \emph{lightstage}. This serves two purposes. On the one hand, spherical gradient illumination provides a means to perform photometric stereo that avoids problems caused by binary shadowing in point source photometric stereo. On the other hand, the use of polarising filters on the lights and camera enables separation of diffuse and specular reflectance which, for the constant illumination case, allows measurement of intrinsic albedo. This was extended to realtime performance capture by Wilson~\etal~\cite{Wilson:10} who showed how a certain sequence of illumination conditions allowed for temporal upsampling of the photometric shape estimates. The main drawback of the lightstage setup is that the required illumination polariser orientation is view dependent and so diffuse/specular separation is only possible for a single viewpoint which does not permit capturing full ear-to-ear face models. Ghosh~\etal~\cite{ghosh2011multiview} made an empirical observation that using two illumination fields with locally orthogonal patterns of polarisation 
allows approximate specular/diffuse separation from any viewpoint on the equator. Although practically useful, in this configuration specular and diffuse reflectance is not fully separated.
More generally, lightstage albedo bakes in ambient occlusion (which depends on geometry) and RGB values are dependent on the light source spectra and camera spectral sensitivities.

\mypara{3D Morphable Model Fitting} The estimation of 3DMM parameters (shape, expression, colour, illumination and camera) is an ongoing inverse rendering challenge. Most approaches focus on shape estimation only and omit the reconstruction of colour/albedo and illumination, e.g.~\cite{liu2018disentangling}. The few methods taking the colour into account suffer from the ambiguity between albedo and illumination demonstrated in Egger~\etal~\cite{egger17phd}. This ambiguity is especially hard to overcome for two reasons: 1.~all publicly available face models don't model real diffuse or specular albedo, 2.~most models have a strong bias towards Caucasian faces which results in a strongly biased prior.
The reflectance models used for inverse rendering are usually dramatically simplified and the specular term is either omitted or constant.  Genova~\etal~\cite{genova2018unsupervised} point out the limitation of no statistics on specularity and use a heuristic for their specular term.  Romdhani~\etal~\cite{Romdhani2005} use the position of specularities as shape cues but again with homogeneous specular maps.
The work of Yamaguchi~\etal~\cite{yamaguchi2018high} demonstrate the value of separate estimation of specular and diffuse albedo, however they do not explore the statistics or build a generative model and their approach is not available to the community.
Current limitations are mainly caused by the lack of a publicly available diffuse and specular albedo model.

\section{Data capture}

A lightstage exploits the phenomenon that specular reflection from a dielectric material preserves the plane of polarisation of linearly polarised incident light whereas subsurface diffuse reflection randomises it. This allows separation of specular and diffuse reflectance by capturing a pair of images under polarised illumination. A polarising filter on each lightsource is oriented such that a specular reflection towards the viewer has the same plane of polarisation. The first image, $I_{\text{para}}$, has a polarising filter in front of the camera oriented parallel to the plane of polarisation of the specularly reflected light, allowing both specular and diffuse transmission. The second, $I_{\text{perp}}$, has the polarising filter oriented perpendicularly, blocking the specular but still permitting transmission of the diffuse reflectance. The difference, $I_{\text{para}}-I_{\text{perp}}$, gives only the specular reflection.

\mypara{Setup} Our setup comprises a custom built lightstage with polarised LED illumination, a single photometric camera (Nikon D200) with optoelectric polarising filter (LC-Tec FPM-L-AR) and seven additional cameras (Canon 7D) to provide multiview coverage. We use 41 ultra bright white LEDs mounted on a geodesic dome of diameter 1.8m. Each LED has a rotatable linear polarising filter in front of it. Their orientation is tuned by placing a sphere of low diffuse albedo and high specular albedo (a black snooker ball) in the centre of the dome and adjusting the filter orientation until the specular reflection is completely cancelled in the photometric camera's view. Since we only seek to estimate albedo maps, we require only the constant illumination condition in which all LEDs are set to maximum brightness. 

\begin{figure*}[!t]
    \centering
    \includegraphics[trim=0px 255px 0px 0px]{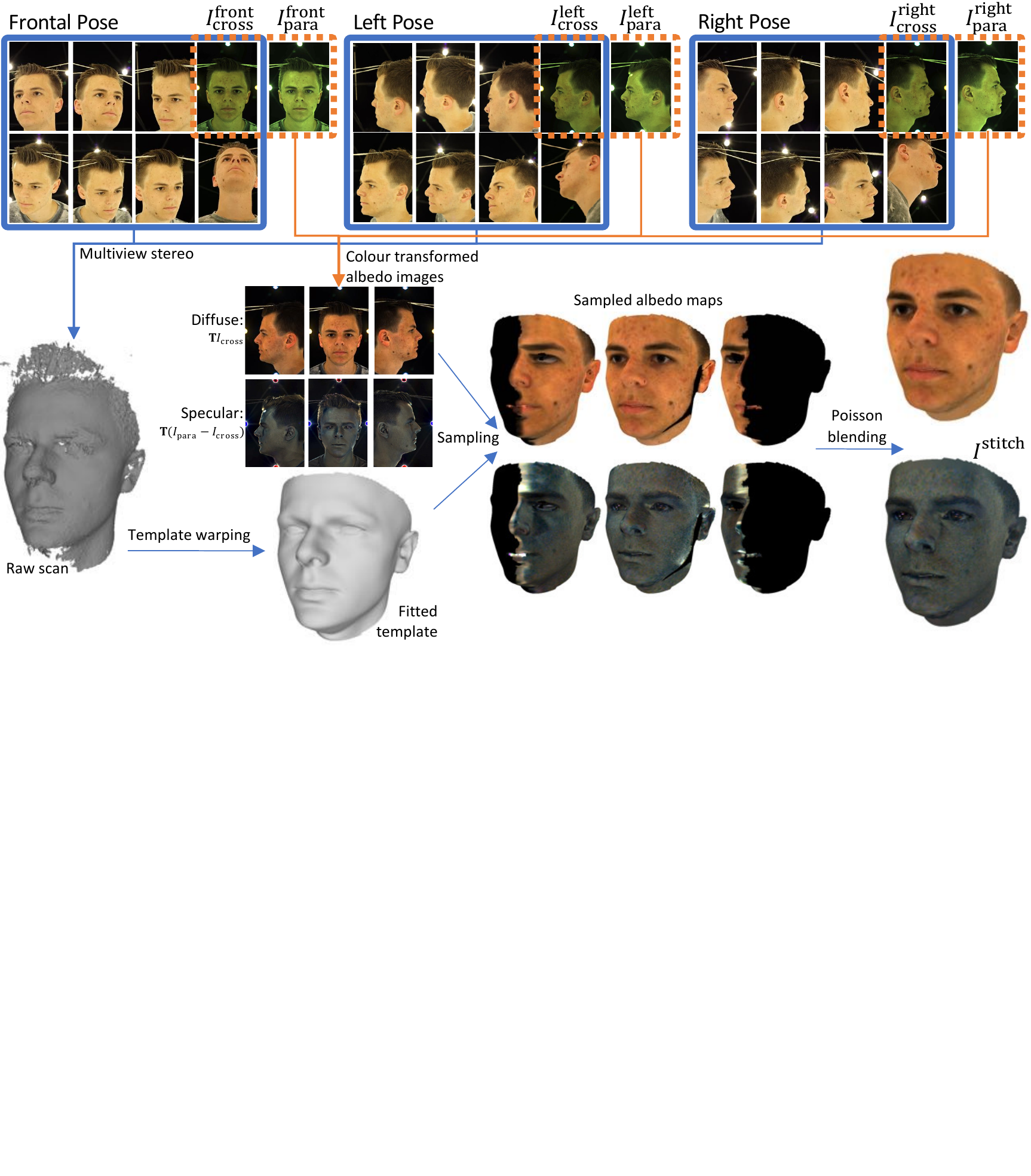}
    \caption{Overview of our capture and blending pipeline. Images within a blue box are captured simultaneously. Photometric image pairs within a dashed orange box are captured sequentially with perpendicular/parallel polarisation state respectively.}
    \label{fig:image_set}
\end{figure*}

In contrast to previous lightstage-based methods, we capture multiple virtual viewpoints by capturing the face in different poses, specifically frontal and left/right profile. This provides full ear-to-ear coverage for the single polarisation-calibrated photometric viewpoint. The optoelectric polarising filter enables the parallel/perpendicular conditions to be captured in rapid succession without requiring mechanical filter rotation. We augment the photometric camera with additional cameras providing multiview, single-shot images captured in sync with the photometric images. We position these additional cameras to provide overlapping coverage of the face. We do not rely on a fixed geometric calibration, so the exact positioning of these cameras is unimportant and we allow the cameras to autofocus between captures. In our setup, we use 7 such cameras in addition to the photometric view giving a total of 8 simultaneous views. Since we repeat the capture three times, we have 24 effective views. For synchronisation, we control camera shutters and the polarisation state of the photometric camera using an MBED micro controller. A complete dataset for a face is shown in Fig.~\ref{fig:image_set}.

\mypara{Participants} We captured 50 individuals (13 females) in our setup. Our participants range in age from 18 to 67 and cover skin types I-V of the Fitzpatrick scale \cite{fitzpatrick1988validity}.

\section{Data processing}

In order to merge these views and to provide a rough base mesh, we perform a multiview reconstruction. We then warp the 3DMM template mesh to the scan geometry. As well as other sources of alignment error, since the three photometric views are not acquired simultaneously, there is likely to be non-rigid deformation of the face between these views. For this reason, in Section \ref{sec:stitch} we propose a robust algorithm for stitching the photometric views without blurring potentially misaligned features. We provide an implementation of our sampling, weighting and blending pipeline as an extension of the MatlabRenderer toolbox \cite{bas2019what}.

\subsection{Multiview stereo}

We commence by applying uncalibrated structure-from-motion followed by dense multiview stereo \cite{agisoft2019agisoft} to all 24 viewpoints (see Fig.~\ref{fig:image_set}, blue boxed images). Solving this uncalibrated multiview reconstruction problem provides both the base mesh (see Fig.~\ref{fig:image_set}, bottom left) to which we fit the 3DMM template and also intrinsic and extrinsic camera parameters for the three photometric views. These form the input to our stitching process.

\subsection{Template fitting}\label{sec:tempfit}
To build a 3DMM from raw scanning data, we establish correspondence to a template. We use the Basel Face Pipeline \cite{gerig2018morphable} which uses smooth deformations based on Gaussian Processes. We adopted the threshold to exclude vertices from the optimisation for the different levels (to 32mm, 16mm, 8mm, 4mm, 2mm, 1mm, 0.5mm from coarse to fine) to reach better performance for missing parts of the scans. Besides this minor change we used the Basel Face Pipeline as is, with between 25 and 45 manually annotated landmarks (eyes: 8, nose 9, mouth 6, eyebrows 4, ears 18). We used the template of the BFM 2017 for registration which makes our model compatible to this model.

\subsection{Sampling and stitching}\label{sec:stitch}

We stitch the multiple photometric viewpoints into seamless diffuse and specular per-vertex albedo maps using Poisson blending. Blending in the gradient domain via solution of a Poisson equation was first proposed by P{\'e}rez~\etal~\cite{Perez:03} for 2D images. 
The approach allows us to avoid visible seams where texture or geometry from different views are inconsistent.

For each viewpoint, $v\in\mathcal{V}=\{v_1,\dots,v_k\}$, we sample RGB intensities onto the $n$ vertices of the mesh, $\mathbf{I}^{v}\in\R^{n\times 3}$. Then, for each view we compute a per-triangle confidence value for each of the $t$ triangles, $\mathbf{w}^v\in\R^{t}$. For each triangle, this is defined as the minimum per-vertex weight for each vertex in the triangle, where the per-vertex weights are defined as follows. If the vertex is not visible in that view, the weight is set to zero. We also set the weight to zero if the vertex projection is within a threshold distance of the occluding boundary to avoid sampling background onto the mesh. Otherwise, we take the dot product between the surface normal and view vectors as the weight, giving preference to observations whose projected resolution is higher.

Next, we define a selection matrix for each view, $\mathbf{S}_{v}\in\{0,1\}^{m_v\times t}$, that selects a triangle if view $v$ has the highest weight for that triangle:
\begin{equation}
\left( \mathbf{S}_{v}^T\mathbf{1}_{m_v} \right)_i = 1 \ \ \text{iff}\ \forall u\in\mathcal{V}\setminus\{v\}, w^u_i<w^v_i. 
\end{equation}
We define an additional selection matrix $\mathbf{S}_{v_{k+1}}$ that selects all triangles not selected in any view (i.e.~that have no non-zero weight). Hence, every triangle is selected exactly once and $\sum_{i=1}^{k+1}m_{v_i}=t$. We similarly define per-vertex selection matrices $\tilde{\mathbf{S}}_{v}\in\{0,1\}^{\tilde{m}_v\times n}$ that select the vertices for which view $v$ has the highest per-vertex weights.

We write a screened Poisson equation as a linear system \cite{dessein2014seamless} in the unknown stitched RGB intensities $\mathbf{I}^{\text{stitch}}\in\R^{n\times 3}$:
\begin{equation}
    \begin{bmatrix}
    \mathbf{SG} \\
    \lambda\tilde{\mathbf{S}}_{v_1}
    \end{bmatrix}
    \mathbf{I}^{\text{stitch}} = 
    \begin{bmatrix}
    (\mathbf{I}_3\otimes\mathbf{S}_{v_1})\mathbf{G}\mathbf{I}^{v_1}\\
    \vdots \\
    (\mathbf{I}_3\otimes\mathbf{S}_{v_k})\mathbf{G}\mathbf{I}^{v_k}\\
    \mathbf{0}_{3m_{k+1}\times 3}\\
    \lambda\tilde{\mathbf{S}}_{v_1}\mathbf{I}^{v_1}
    \end{bmatrix},\label{eqn:poisson}
\end{equation}
where $\otimes$ is the Kronecker product,
\begin{equation}
\mathbf{S}=
    \begin{bmatrix}
    \mathbf{I}_3\otimes\mathbf{S}_{v_1}\\
    \vdots\\
    \mathbf{I}_3\otimes\mathbf{S}_{v_{k+1}}\\
    \end{bmatrix},
\end{equation}
$\mathbf{I}_3$ is the $3\times 3$ identity matrix and $\mathbf{G}\in\R^{3t\times n}$ computes the per-triangle gradient in the $x$, $y$ and $z$ directions of a function defined on the $n$ vertices of the mesh. We solve \eqref{eqn:poisson} in a least squares sense so that $\mathbf{I}^{\text{stitch}}$ seeks to match the selected gradients in each triangle. Triangles with no selected view are assumed to have zero gradient. View $v_1$ is chosen as the reference in order to resolve colour offset indeterminacies and $\lambda$ is the screening weight. We use $k=3$ views, the frontal view is chosen as the reference and we set $\lambda=0.1$.

\subsection{Calibrated colour transformation}

Our photometric camera captures RAW linear images. We transform these to linear sRGB space using a colour transformation matrix computed from light SPD and camera spectral sensitivity calibrations, discretised at $D$ evenly spaced wavelengths. We measure the spectral power distribution of the LEDs used in our lightstage, $\mathbf{e}\in\R^D$, using a B\&W Tek BSR111E-VIS spectroradiometer. We use the spectral sensitivity measurement, $\mathbf{C}\in\R^{D\times 3}$, for the Nikon D200 as included in the database of Jiang~\etal~\cite{jiang2013space}. The overall colour transformation is given by a product of three transformations: $\mathbf{T}=\mathbf{T}_{\textrm{xyz2rgb}}\mathbf{T}_{\textrm{raw2xyz}}(\mathbf{C})\mathbf{T}_{\textrm{wb}}(\mathbf{C},\mathbf{e})$. The first performs white balancing:
\begin{equation}
    \mathbf{T}_{\textrm{wb}}(\mathbf{C},\mathbf{e}) = \textrm{diag}(\mathbf{C}^T\mathbf{e})^{-1}.
\end{equation}
The second converts from the camera-specific colour space to the standardised XYZ space:
\begin{equation}
    \mathbf{T}_{\textrm{raw2xyz}}(\mathbf{C}) = \mathbf{C}_{\text{CIE}}\mathbf{C}^+,
\end{equation}
where $\mathbf{C}_{\text{CIE}}\in\R^{D\times 3}$ contains the wavelength discrete CIE-1931 2-degree color matching function and $\mathbf{C}^+$ is the pseudoinverse of $\mathbf{C}$. To preserve white balance we rescale each row such that: $\mathbf{T}_{\textrm{raw2xyz}}(\mathbf{C})\mathbf{1}=\mathbf{1}$. The final transformation, $\mathbf{T}_{\textrm{xyz2rgb}}$, is a fixed matrix to convert from XYZ to sRGB space. As part of our model we provide $\mathbf{T}$, $\mathbf{C}$ and $\mathbf{e}$.

\section{Integrating 3DRFE}

\begin{figure}
    \centering
    \includegraphics[width=\columnwidth]{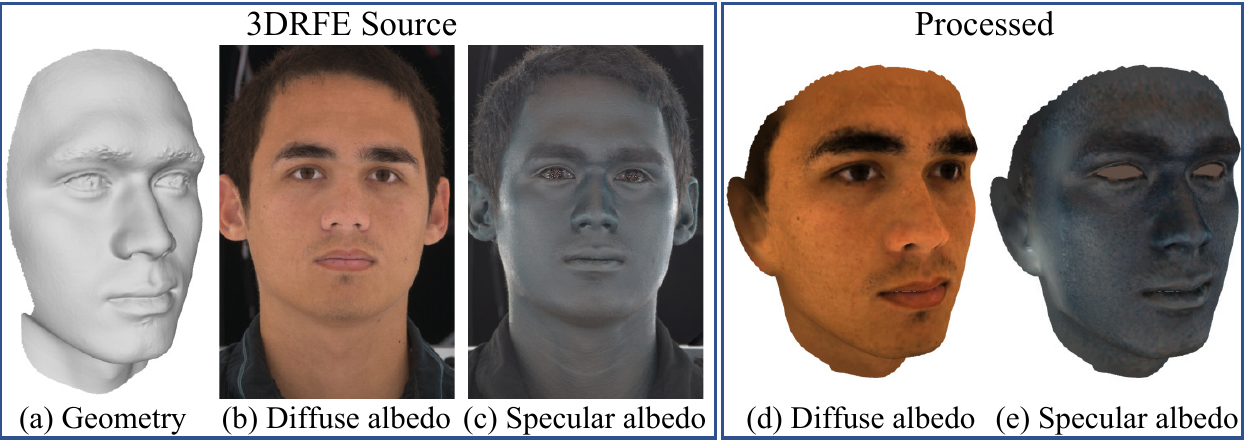}
    \caption{(a)-(c): Source geometry and albedo maps from the 3DRFE dataset \cite{stratou2011effect}. (d)-(e): final registered, colour transformed albedo maps on warped template geometry.}
    \label{fig:3DRFE}
\end{figure}

We augment our own dataset by additionally including the 23 scans from the 3DRFE dataset \cite{stratou2011effect}. This uses the original capture setup of Ma~\etal~\cite{ma2007rapid} which means that photometric information is only captured from the one view for which the polariser orientations are calibrated. Scans are provided in the form of single viewpoint specular and diffuse albedo maps and a mesh (see Fig.~\ref{fig:3DRFE}(a)-(c)) whose UV coordinates are the 2D perspective projection of the mesh into the maps. This enables us to estimate geometric camera calibration parameters from the 3D vertex positions and corresponding 2D UV coordinates. We perform the calibration using \cite{bouguet2008camera} and estimate both intrinsics and distortion parameters. We fit the BFM template to the meshes in the same way as for our own data (see Section \ref{sec:tempfit}). We then project the fitted template into the maps using the estimated camera calibration, directly sample diffuse/specular albedo for visible vertices and inpaint vertices with no sample using a zero gradient assumption.

The diffuse and specular albedo maps are stored in a nonlinear colour space so we preprocess them by applying inverse gamma (of value $2.2$) to transform them back to a linear space. To account for variation in overall skin brightness, during capture the camera gain (ISO) was adjusted for each subject. This means that albedo maps cannot be directly compared or modelled since their individual scale is different. We obtained from the original authors the ISO setting for each subject and compensate by dividing each albedo map by its ISO number. Finally, the albedo maps differ from those taken in our setup by an unknown overall scale factor and colour transformation. To compensate for this, we find the optimal $3\times 3$ colour transformation to transform the mean diffuse albedo of the 3DRFE scans onto the mean of our scans. We apply this transformation to all of the linearised, ISO-normalised albedo maps to give the final set of maps used in our model.

\section{Modelling}

We model diffuse and specular albedo using a linear statistical model learnt with PCA:
\begin{equation}
    \mathbf{x}(\mathbf{b})=\mathbf{Pb}+\bar{\mathbf{x}}, \label{eqn:linmodel}
\end{equation}
where $\mathbf{P}\in\R^{3n\times d}$ contains the $d$ principal components, $\bar{\mathbf{x}}\in\R^{3n}$ is the vectorised average map and $\mathbf{x}:\R^d\mapsto \R^{3n}$ is the generator function that maps from the low dimensional parameter vector $\mathbf{b}\in\R^d$ to a vectorised albedo map. Whilst there are more elaborate techniques to model facial texture, we decided to use PCA because of its very stable performance even in the very low data regime and its quality in terms of generalisation and specificity.

\mypara{Inpainting} The stitched albedo maps produced by the process described in Section \ref{sec:stitch} may still contain artefacts, for example in regions with no observed data, stray hairs across the face, where background is sampled onto the face or due to alignment errors in the pipeline. In addition, some faces in the 3DRFE database have closed eyes which is not desired in our model. For this reason, we manually mask all regions containing artefacts (amounting to 5\% of the vertices in our dataset) and complete them using a novel hybrid of statistical inpainting and Poisson blending. 

For each sample, we assume the set $\mathcal{M}\subset\{1,\dots,n\}$ contains a subset of the $n$ model vertices that have been masked out. We begin by computing a linear statistical model \eqref{eqn:linmodel} in which masked out values are replaced by the average over non-missing values. 

\begin{figure}
    \centering
    \includegraphics[width=\columnwidth]{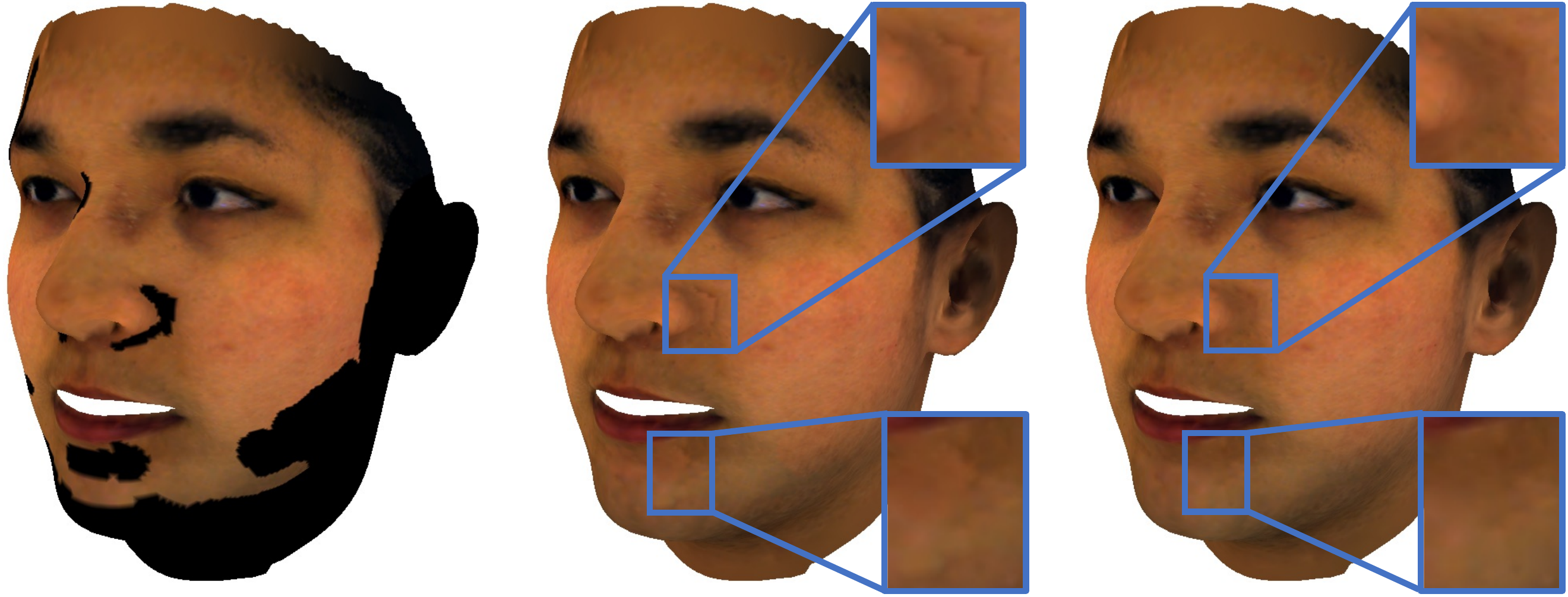}
    \caption{Hole filling (subject with most masked vertices). Left: manually masked albedo map. Middle: statistically inpainted. Right: Poisson blend.}
    \label{fig:inpainting}
\end{figure}

As before, we define a selection matrix for the masked and non-masked vertices, $\tilde{\mathbf{S}}_{\mathcal{M}}$ and $\tilde{\mathbf{S}}_{\mathcal{M}^{\prime}}$ respectively. We also define selection matrices for the triangles whose vertices are all masked, $\mathbf{S}_{\mathcal{M}}$, all non-masked, $\mathbf{S}_{\mathcal{M}^{\prime}}$, and the $s$ triangles that contain a mix of masked and non-masked vertices $\mathbf{S}_{\text{mix}}\in\R^{s\times t}$. We compute the parameters of a least squares fit of the preliminary model to the stitched colours of the non-masked vertices:
\begin{equation}
    \mathbf{b}^* = ((\mathbf{1}_3\otimes\tilde{\mathbf{S}}_{\mathcal{M}^{\prime}})\mathbf{P})^+(\mathbf{1}_3\otimes\tilde{\mathbf{S}}_{\mathcal{M}^{\prime}})(\text{vec}(\mathbf{I}^{\text{stitch}})-\bar{\mathbf{x}}),
\end{equation}
where $^+$ denotes the pseudoinverse. We compute the final albedo maps by again writing a screened Poisson equation as a linear system:
\begin{equation}
    \begin{bmatrix}
    
    (\mathbf{1}_3\otimes\mathbf{S}_{\mathcal{M}})\mathbf{G}\\
    (\mathbf{1}_3\otimes\mathbf{S}_{\text{mix}})\mathbf{G}\\
    \tilde{\mathbf{S}}_{\mathcal{M}^{\prime}}\\
    
    \end{bmatrix}
    \mathbf{I}^{\text{complete}} = 
    \begin{bmatrix}
    
    (\mathbf{1}_3\otimes\mathbf{S}_{\mathcal{M}})\mathbf{G}\mathbf{I}^{\text{stat}} \\
    \mathbf{0}_{s}\\
    \tilde{\mathbf{S}}_{\mathcal{M}^{\prime}}\mathbf{I}^{\text{stitch}}\\
    
    \end{bmatrix}
\end{equation}
where $\text{vec}(\mathbf{I}^{\text{stat}})=\mathbf{P}\mathbf{b}^*+\bar{\mathbf{x}}$ is the statistically inpainted texture. The solution encourages the texture gradient in the masked out region to match the gradient of the statistically inpainted texture but to match the original texture in the non-masked region. For triangles on the boundary between masked and non-masked regions we encourage zero gradient. In Fig.~\ref{fig:inpainting} we show an example for the face with most masking required. Note that simply using the statistical inpainting (middle) leads to seams in the texture. The process can be iterated so that these completed textures are used to rebuild the statistical model though we note no significant improvement after the first iteration. We apply this masking and blending procedure to both diffuse and specular albedo maps. 

We perform an additional final step for the specular maps. Specular albedo is not meaningfully estimated in the eyeball region. This is because the eyeball surface is highly specular compared to skin (i.e.~the specular lobe is much narrower). Since the spherical illumination is discretised by a relatively small number of light sources, most points on the eye surface do not specularly reflect towards the viewer (see Fig.~\ref{fig:3DRFE}(c) - zoom for detail). For this reason, we replace specular albedo values in the eyeball region by a robust maximum (95th percentile) of the estimated specular albedo values in that region (see Fig.~\ref{fig:3DRFE}(e)).

\begin{figure}
    \centering
    \includegraphics[width=\columnwidth]{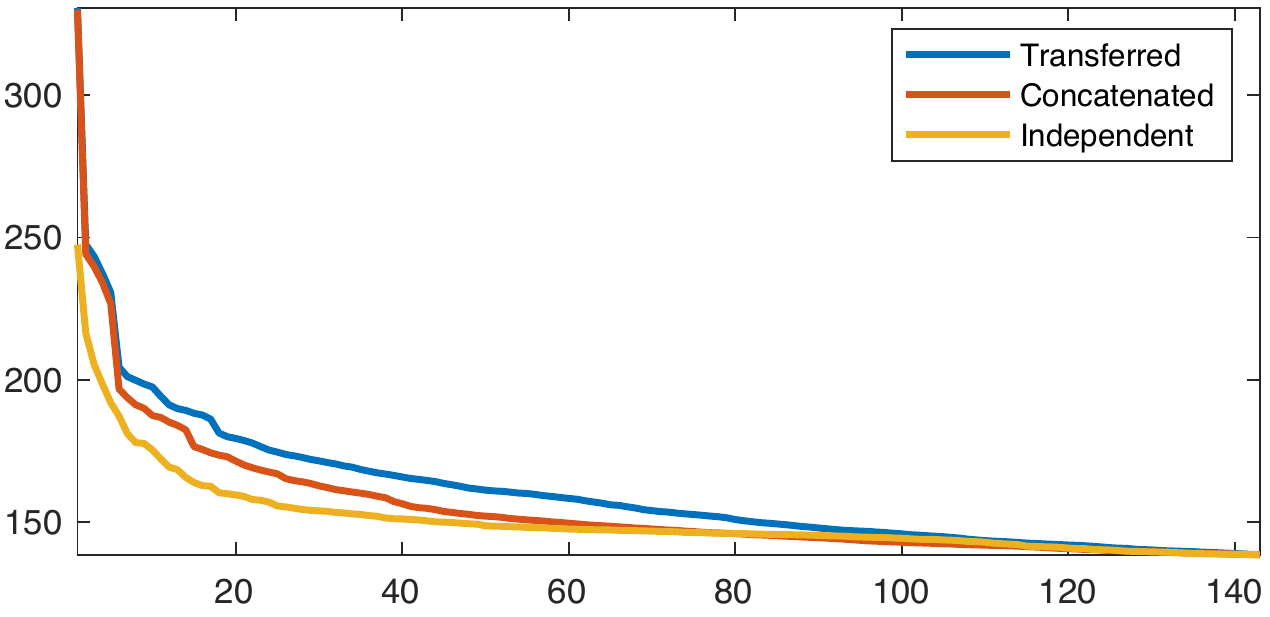}
    \caption{Leave-one-out generalisation error for three variants of the specular model.}
    \label{fig:specgen}
\end{figure}

\mypara{Statistical modelling} The most straightforward way to model diffuse and specular albedo is with two separate models of the same form as equation \eqref{eqn:linmodel}. However, a drawback of this is that the two maps are not independent and allowing arbitrary combinations of the two model parameters can lead to unrealistic appearance. For example, if the face has a beard in the diffuse albedo map, then the specular albedo should be lower in the beard region. An obvious alternative is to learn a joint model in which diffuse and specular are concatenated and modelled together. A drawback of this model is that it may be desirable to retain different numbers of principal components for the two models or to use the diffuse model alone. Using only the diffuse part of this joint model is no longer orthonormal. In addition, since diffuse albedo conveys most of the information about the identity of a face, it is desirable to have the statistics focused on the diffuse part. For these reasons, we propose an additional third alternative. Here, we learn a diffuse only model and then build a specular model in which the principal components are made from the same linear combinations of training samples as the diffuse modes. This means that the same parameters can be used for both models while retaining orthonormality of the diffuse model:
\begin{align}
    \text{vec}(\mathbf{I}^{\text{diff}}) &= \mathbf{P}^{\text{diff}}\mathbf{b}+\bar{\mathbf{x}}^{\text{diff}},\\
    \text{vec}(\mathbf{I}^{\text{spec}}) &= \mathbf{P}^{\text{spec}}\mathbf{b}+\bar{\mathbf{x}}^{\text{spec}},
\end{align}
Comparing the three alternatives (see Fig.~\ref{fig:specgen}), the independent specular model generalises best, the concatenated model second best and the proposed model with principal component weights transferred from the diffuse model worst. However, the difference is neglectable and the combination of having a single set of parameters for both models and retaining optimality of the independent diffuse model makes this the best choice.

We use symmetry augmentation in our modelling. The BFM template is bilaterally symmetric with known symmetry correspondences. Therefore, we include each sample twice, once as captured, once reflected. This gives us a total of 146 training samples.
We make all variants of our model publicly available using both the full BFM 2017 template and also a template cropped only to the inner face region.

\mypara{Image formation model} To use our model for synthesis or analysis-by-synthesis requires a slightly different image formation model than is typically used with 3DMMs. Appearance at a vertex $v$ should be computed as follows:
\begin{equation}
    \mathbf{i}_v = \left[ 
    \mathbf{i}_{\text{diff}}(\mathbf{P}^{\text{diff}}_v\mathbf{b}+\bar{\mathbf{x}}^{\text{diff}}_v) +
    \mathbf{i}_{\text{spec}}(\mathbf{P}^{\text{spec}}_v\mathbf{b}+\bar{\mathbf{x}}^{\text{spec}}_v)
    \right]^{\frac{1}{2.2}}
\end{equation}
where $\mathbf{i}_{\text{diff}}$ and $\mathbf{i}_{\text{spec}}$ are colour diffuse and specular shading (computed using a chosen reflectance model and dependent on illumination, geometry and viewing direction), $\mathbf{P}^{\text{diffuse}}_v$ denotes the three rows of $\mathbf{P}^{\text{diffuse}}$ corresponding to vertex $v$, similarly for $\mathbf{P}^{\text{spec}}_v$, $\bar{\mathbf{x}}^{\text{diff}}_v$ and $\bar{\mathbf{x}}^{\text{spec}}_v$. See Fig.~\ref{fig:teaser} (right) for a visualisation using this image formation model. In addition, for a camera that does not work in sRGB colour space, an additional transformation to the camera's colour space prior to nonlinear gamma of 2.2 is required. 

\section{Experiments}

Our final model is a combination of the proposed diffuse and specular albedo model to model facial appearance and the BFM 2017 to model face shape and expressions. Since the shape part of the model is identical to the BFM 2017, we focus on the evaluation of the appearance model and reconstruction of facial albedo.

\begin{figure}[!t]
    \centering
\begingroup
\setlength{\tabcolsep}{1pt}
\renewcommand{\arraystretch}{0.5}
\begin{tabular}{m{0.3cm}m{1.2cm}c}
 \rotatebox{90}{Ours (diffuse)} &
    \includegraphics[width=1.2cm]{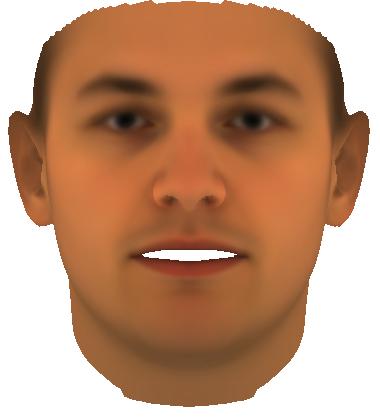} &
    \begin{tabular}{ccccc}
    \includegraphics[width=1.2cm]{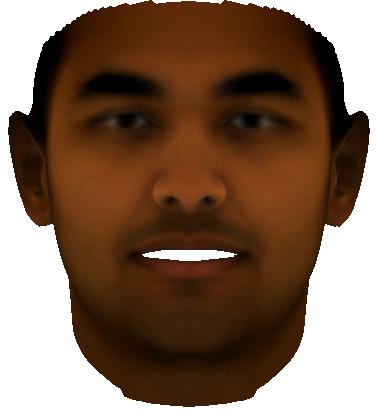} &
    \includegraphics[width=1.2cm]{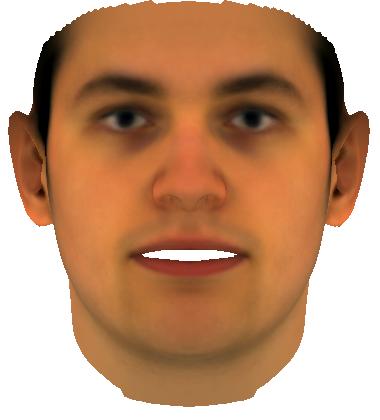} &
    \includegraphics[width=1.2cm]{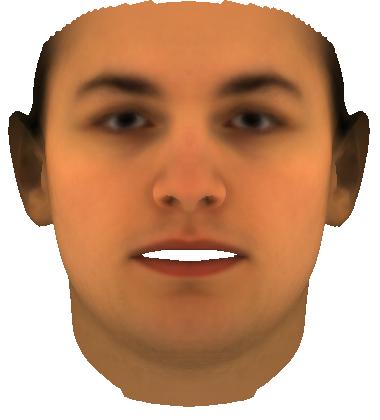} &
    \includegraphics[width=1.2cm]{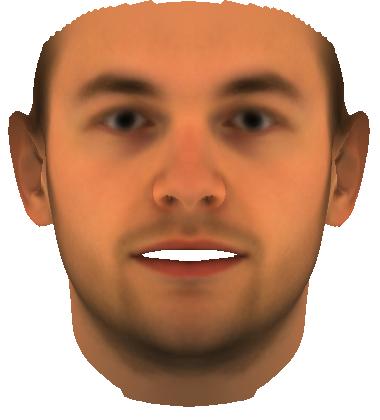} &
    \includegraphics[width=1.2cm]{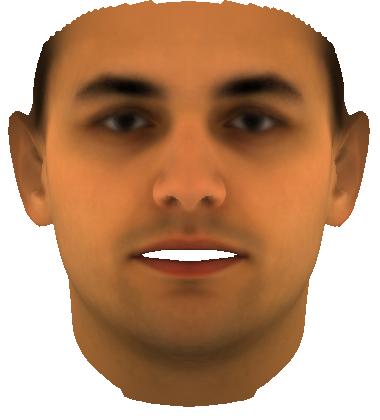} \\
    \includegraphics[width=1.2cm]{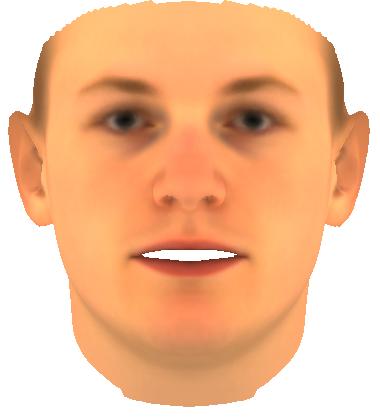} &
    \includegraphics[width=1.2cm]{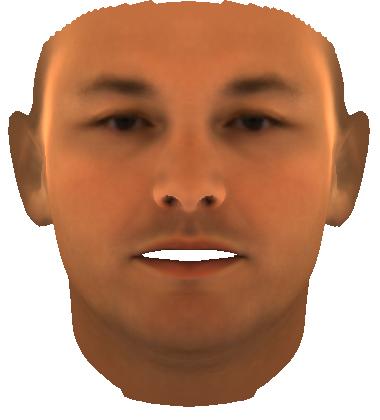} &
    \includegraphics[width=1.2cm]{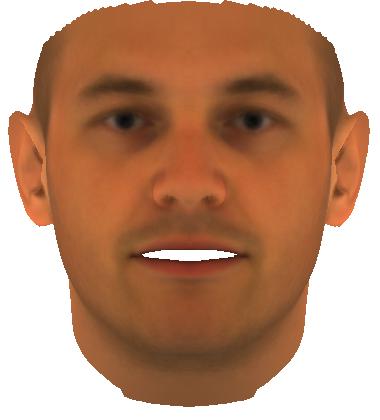} &
    \includegraphics[width=1.2cm]{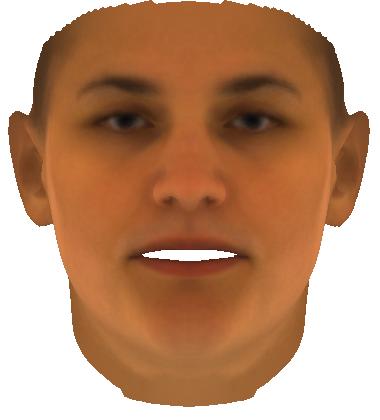} &
    \includegraphics[width=1.2cm]{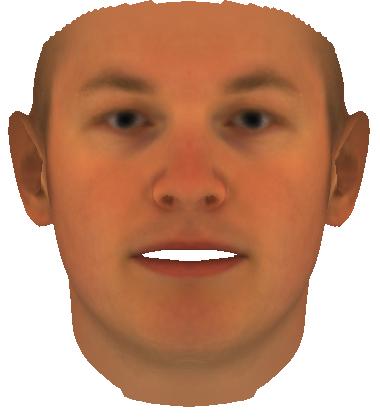} 
    \end{tabular} \vspace{0.1cm} \\
    \hline \\
 \rotatebox{90}{BFM 2017 \cite{gerig2018morphable}} &
    \includegraphics[width=1.2cm]{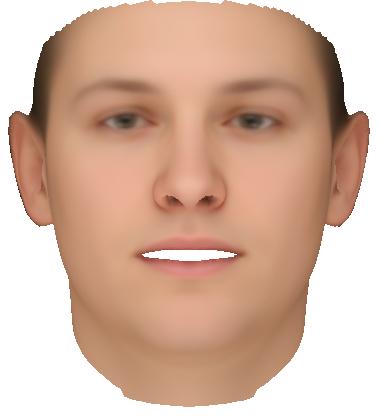} &
    \begin{tabular}{ccccc}
    \includegraphics[width=1.2cm]{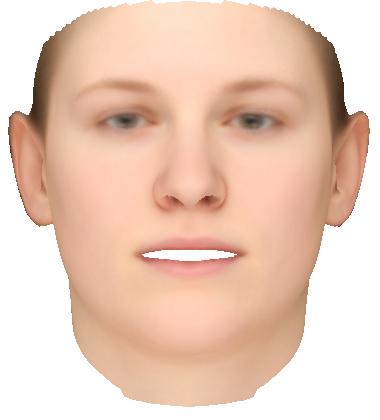} &
    \includegraphics[width=1.2cm]{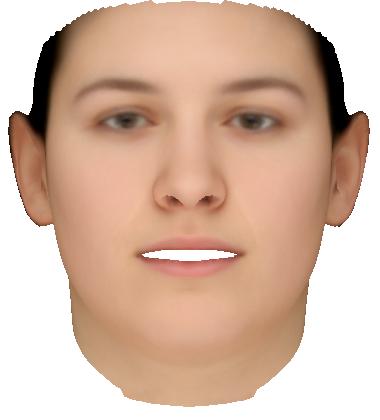} &
    \includegraphics[width=1.2cm]{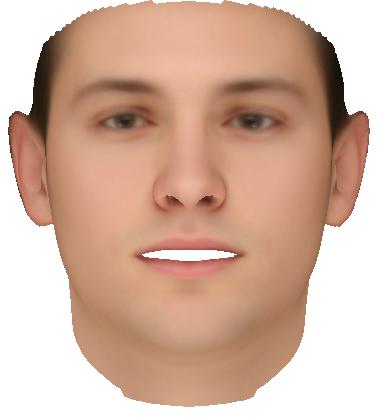} &
    \includegraphics[width=1.2cm]{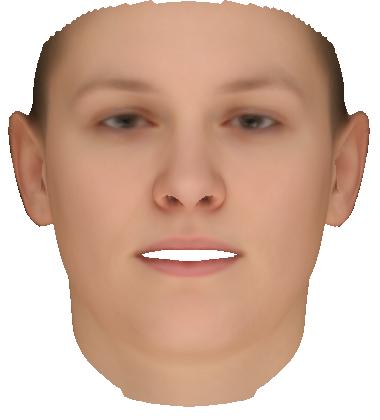} &
    \includegraphics[width=1.2cm]{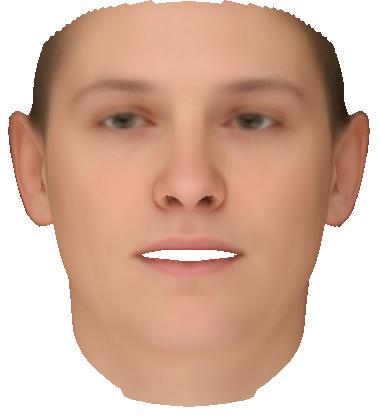} \\
    \includegraphics[width=1.2cm]{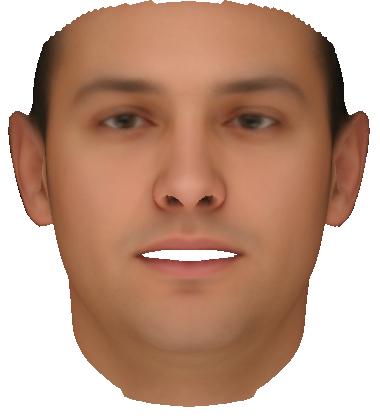} &
    \includegraphics[width=1.2cm]{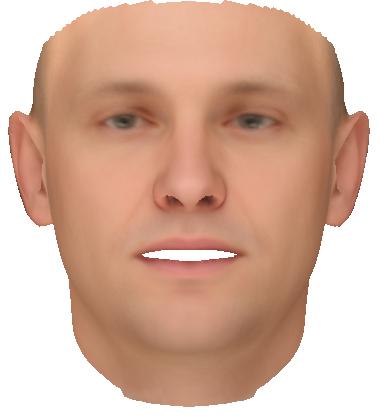} &
    \includegraphics[width=1.2cm]{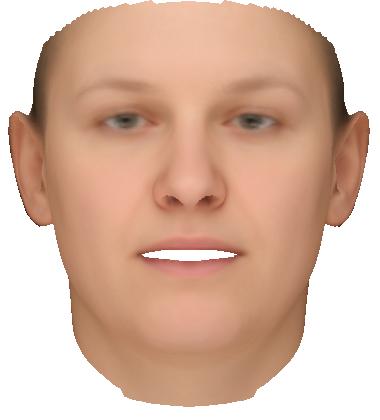} &
    \includegraphics[width=1.2cm]{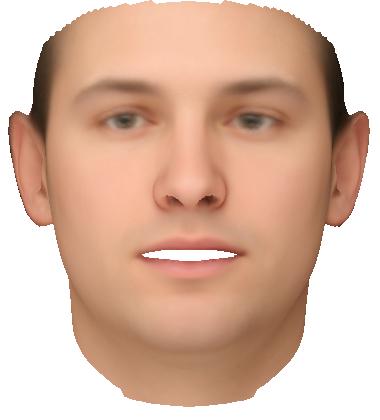} &
    \includegraphics[width=1.2cm]{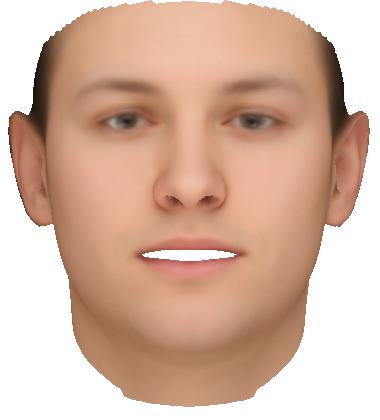} 
    \end{tabular} \vspace{0.1cm} \\
    \hline \\
 \rotatebox{90}{LYHM \cite{LYHM2017}} &
    \includegraphics[width=1.2cm]{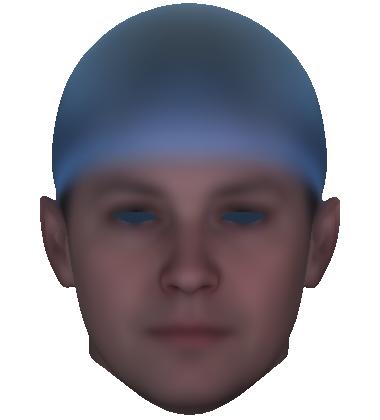} &
    \begin{tabular}{ccccc}
    \includegraphics[width=1.2cm]{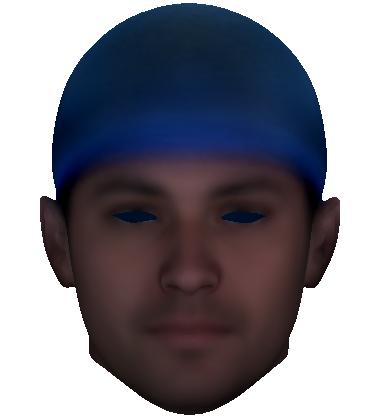} &
    \includegraphics[width=1.2cm]{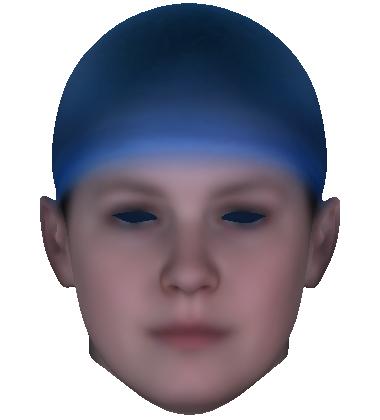} &
    \includegraphics[width=1.2cm]{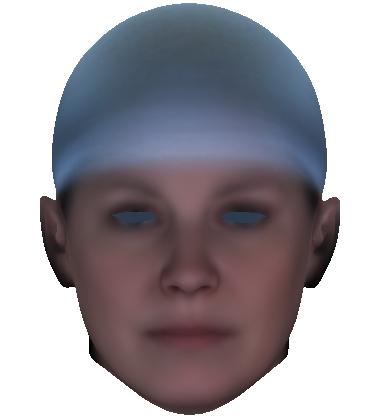} &
    \includegraphics[width=1.2cm]{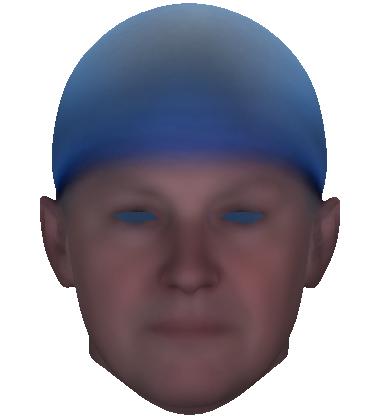} &
    \includegraphics[width=1.2cm]{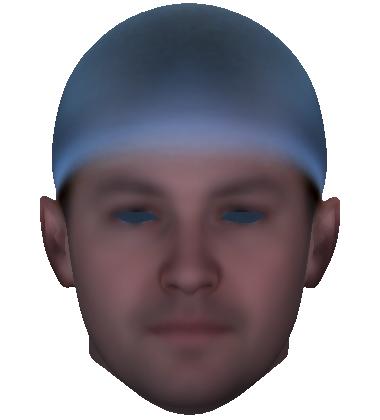} \\
    \includegraphics[width=1.2cm]{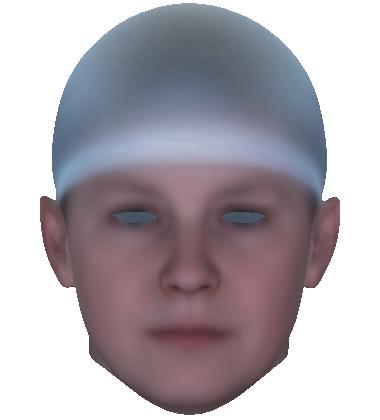} &
    \includegraphics[width=1.2cm]{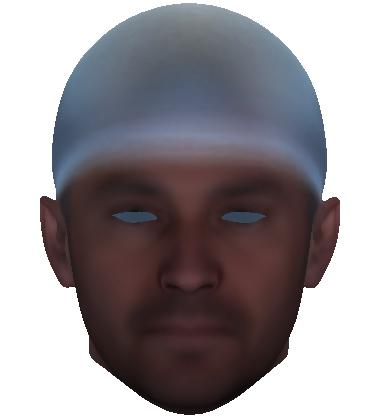} &
    \includegraphics[width=1.2cm]{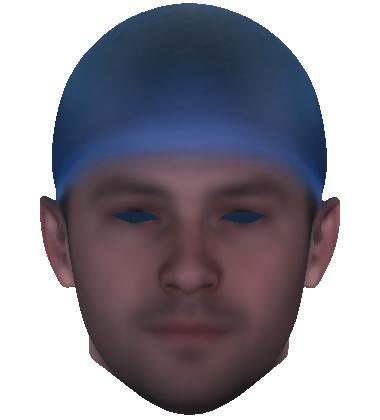} &
    \includegraphics[width=1.2cm]{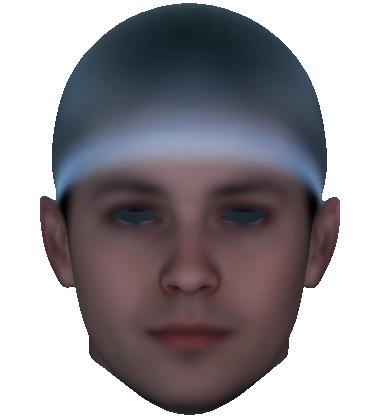} &
    \includegraphics[width=1.2cm]{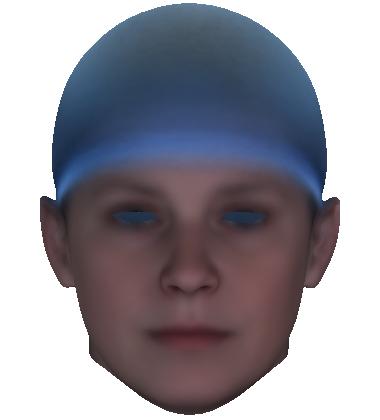} 
    \end{tabular}
\end{tabular}
\endgroup
    \caption{Comparison with current state-of-the-art and publicly available models. Our full model is shown in Fig.~\ref{fig:teaser}}
    \label{fig:modelcomp}
\end{figure}

We begin by providing a qualitative comparison between our proposed model and the currently most used publicly available 3DMMs in Fig.~\ref{fig:modelcomp}. We observe that the first mode of our proposed model is more diverse and less biased than the BFM. Additionally, we see that the appearance between models varies dramatically which shows how arbitrary the albedo in the LYHM and BFM are. Our full model presented in Fig.~\ref{fig:teaser} is unprecedented and there is no other model to compare to.

\begin{figure*}
    \centering
    \includegraphics[width=\textwidth]{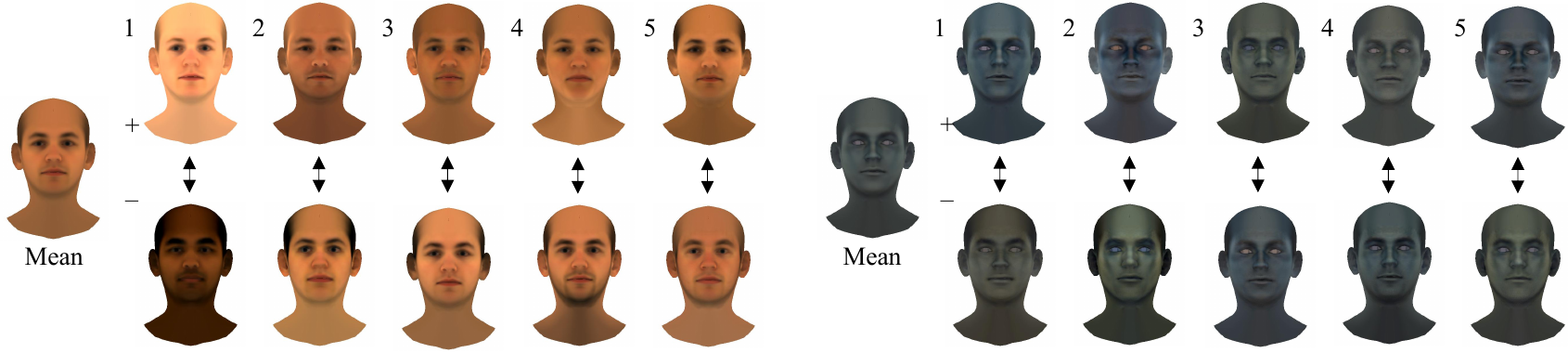}
    \caption{Alternate version of our model in the FLAME \cite{FLAME:SiggraphAsia2017} topology.}
    \label{fig:FLAME}
\end{figure*}

We additionally provide a version of our model in the FLAME \cite{FLAME:SiggraphAsia2017} topology. See Figure \ref{fig:FLAME}. Missing data on this full head model is smoothly extrapolated by assuming zero gradient in those regions.

\begin{figure}[!t]
    \centering
    \includegraphics[width=\columnwidth]{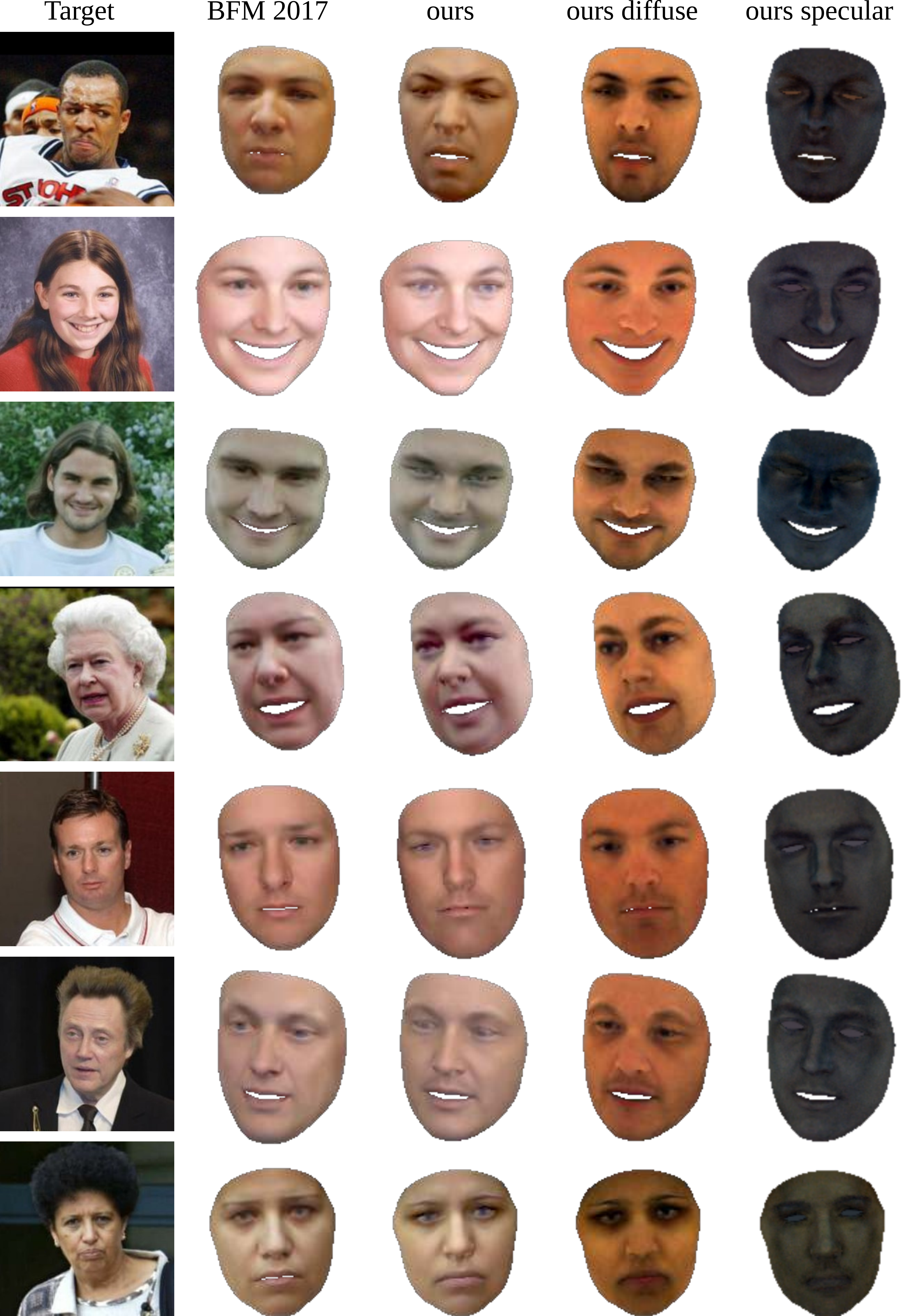}
    \caption{Qualitative model adaptation results on the LFW dataset \cite{LFWTech}. Our model leads to comparable results whilst explicitly disentangling albedo and estimating diffuse and specular albedo.}
    \label{fig:qualitative}
\end{figure}

Next, we use our model in a standard inverse rendering setting. We adopted the publicly available model adaptation framework\footnote{\url{https://github.com/unibas-gravis/basel-face-pipeline}} based on \cite{Schonborn2017} and compare it directly to model adaptation results based on the BFM in Fig.\ref{fig:qualitative}. This implementation adapts shape, albedo and camera parameters, as well as the first three bands of a spherical harmonics illumination model and is based on Markov Chain Monte Carlo Sampling. We perform the experiment on the LFW dataset \cite{LFWTech} exactly as proposed in \cite{gerig2018morphable} and just exchanged the model (including applying gamma) and used statistical specular albedo maps during model adaptation.

\begin{figure}[!t]
    \centering
    \includegraphics[width=\columnwidth]{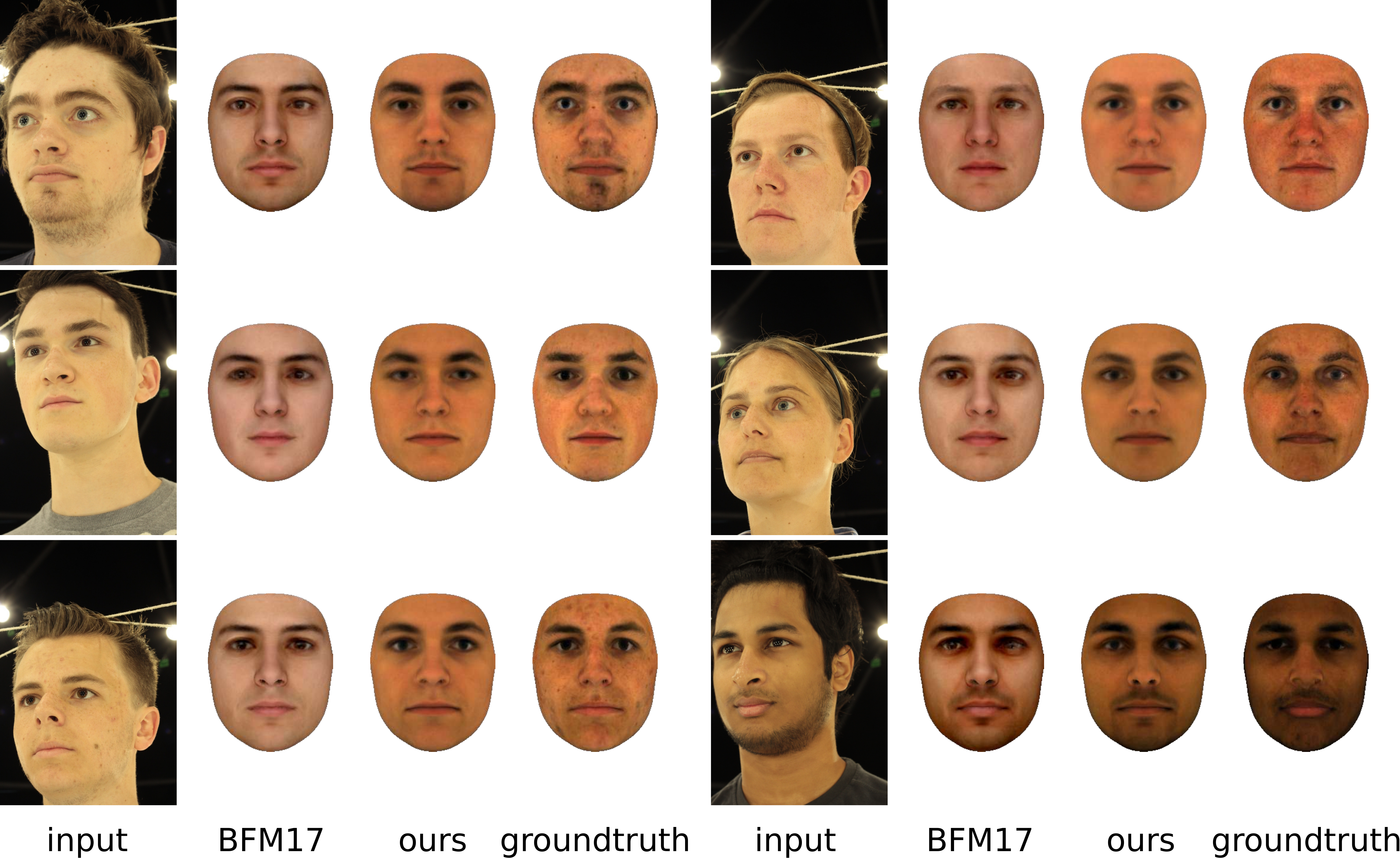}
    \caption{Albedo estimation results based on the exact same inverse rendering pipeline for the BFM 2017 and the proposed model. The proposed model is both visually and in terms of mean squared error (see Table~\ref{tab:benchmark}) closer to the ground truth.}
    \label{fig:benchmark}
\end{figure}

\begin{table}[]
\begin{tabular}{|l|r|l|}
\hline
      & reconstruction  & model mean            \\ \hline
BFM17 & 0.0192 $\pm$ 0.0121  & 0.0575 $\pm$ 0.0551 \\ \hline
ours & \textbf{0.0060 $\pm$ 0.0022} & \textbf{0.0170 $\pm$ 0.0270} \\ \hline
\end{tabular}
\caption{Albedo estimation results for the BFM 2017 and the proposed method. The second column shows the reconstruction based on the respective model mean solely. Those results are based on the reconstructions depicted in Fig~\ref{fig:benchmark}.}
\label{tab:benchmark}
\end{table}

Finally, we perform an evaluation in the same inverse rendering setting as the previous experiment but with known ground truth albedo maps. We use six identities from our own dataset and build a model excluding them. We then fit to images from our dataset taken by the non-photometric cameras. These are simply SLR cameras in auto mode with no polarisation, representing a realistic image in approximately ambient light. We apply the inverse rendering framework with the same configuration, except for limiting the illumination condition to an ambient one and estimate albedo and observe better albedo reconstruction performance for our proposed model compared to the BFM for every single case. We applied gamma for both models since it leads to better results even for the albedo reconstruction of the BFM. Visual results can be found in Fig.~\ref{fig:benchmark} and quantitative values are shown in Table~\ref{tab:benchmark}.


\section{Conclusion}

We built and make available the first statistical model of facial diffuse and specular albedo. The model at hand fills a gap in 3DMM literature and might be beneficial in various directions. This model leverages the computer graphics part of the inverse rendering setting where 3DMMs are classically applied. We present superior performance compared to the BFM 2017 in terms of albedo reconstruction from the facial appearance in a 2D image. Besides the computer vision application of inverse rendering with all its various approaches, we see big potential in the direction of de-biasing current face processing pipelines. To the best of our knowledge, this work is the first to combine diffuse and specular albedo and jointly model different skin types with their matching specular reflection properties. Besides applications for computer graphics and vision, we also see a benefit for studying human face perception. Whilst other 3DMMs were already used in behavioural experiments, this is the first model enabling to study human face perception based on a real disentangled representation of illumination, shading, and reflection. We make our model and accompanying code publicly available\footnote{\url{https://github.com/waps101/AlbedoMM}}.

\section*{Acknowledgement} 

W. Smith is supported by a Royal Academy of Engineering/The Leverhulme Trust Senior Research Fellowship. B. Egger and J. Tenenbaum are supported by the Center for Brains, Minds and Machines (CBMM), funded by NSF STC award CCF-1231216. We acknowledge Abhishek Dutta for the original design and construction of our light stage. We thank Timo Bolkart for registering our data with the FLAME topology.

\bibliographystyle{ieee_fullname}
\bibliography{refs}

\clearpage

\begin{appendices}

\section{Lightstage setup}

\begin{figure}[!b]
    \centering
    \includegraphics[width=\columnwidth]{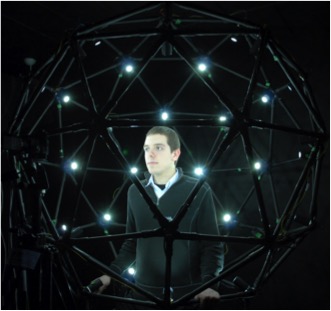}
    \caption{A subject in our lightstage which comprises white LEDs mounted on a once-subdivided icosahedron.}
    \label{fig:lightstage}
\end{figure}

\begin{figure*}
    \centering
    \includegraphics[trim=247px 72px 228px 31px,clip=true,width=5cm]{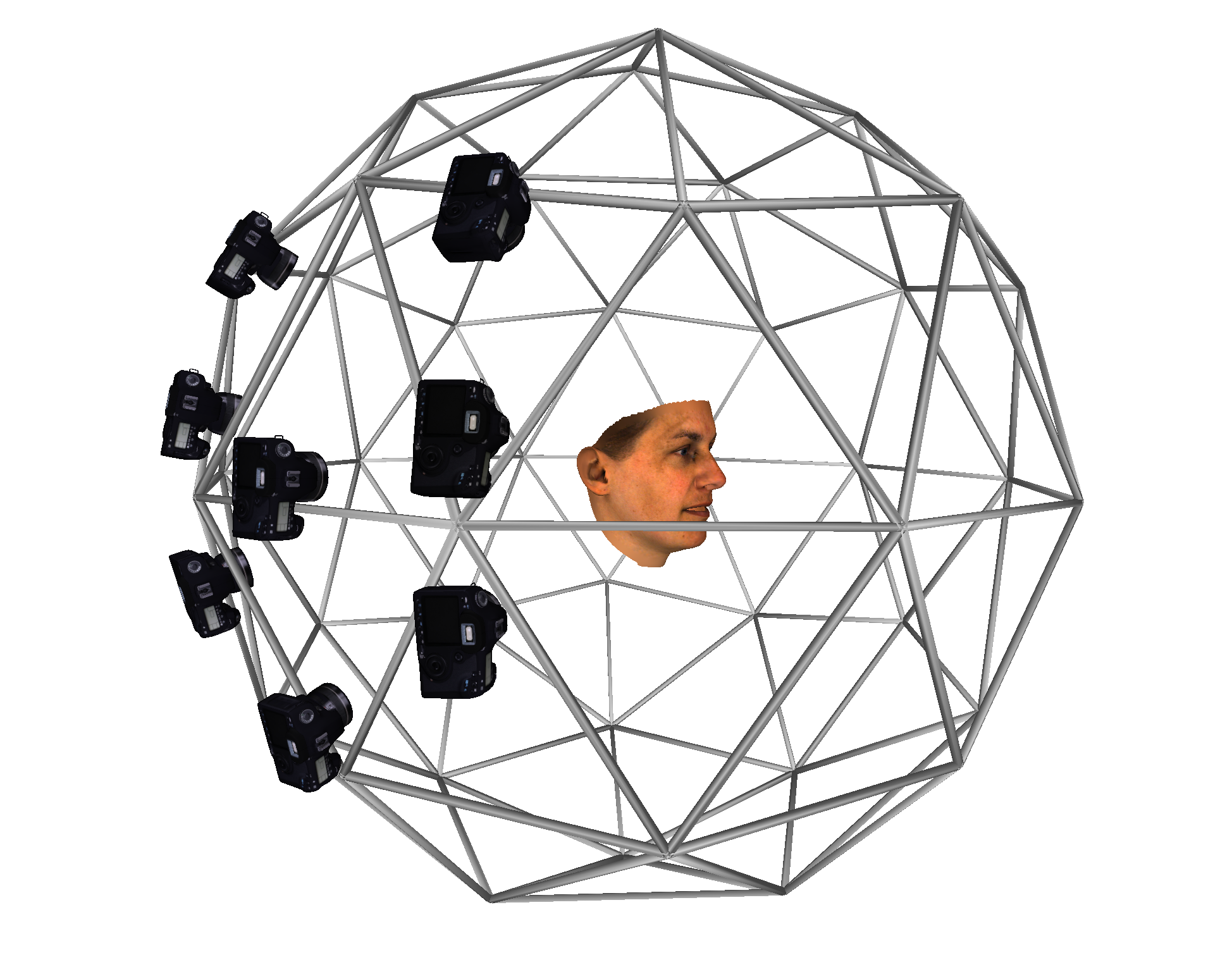}
    \includegraphics[trim=247px 72px 228px 31px,clip=true,width=5cm]{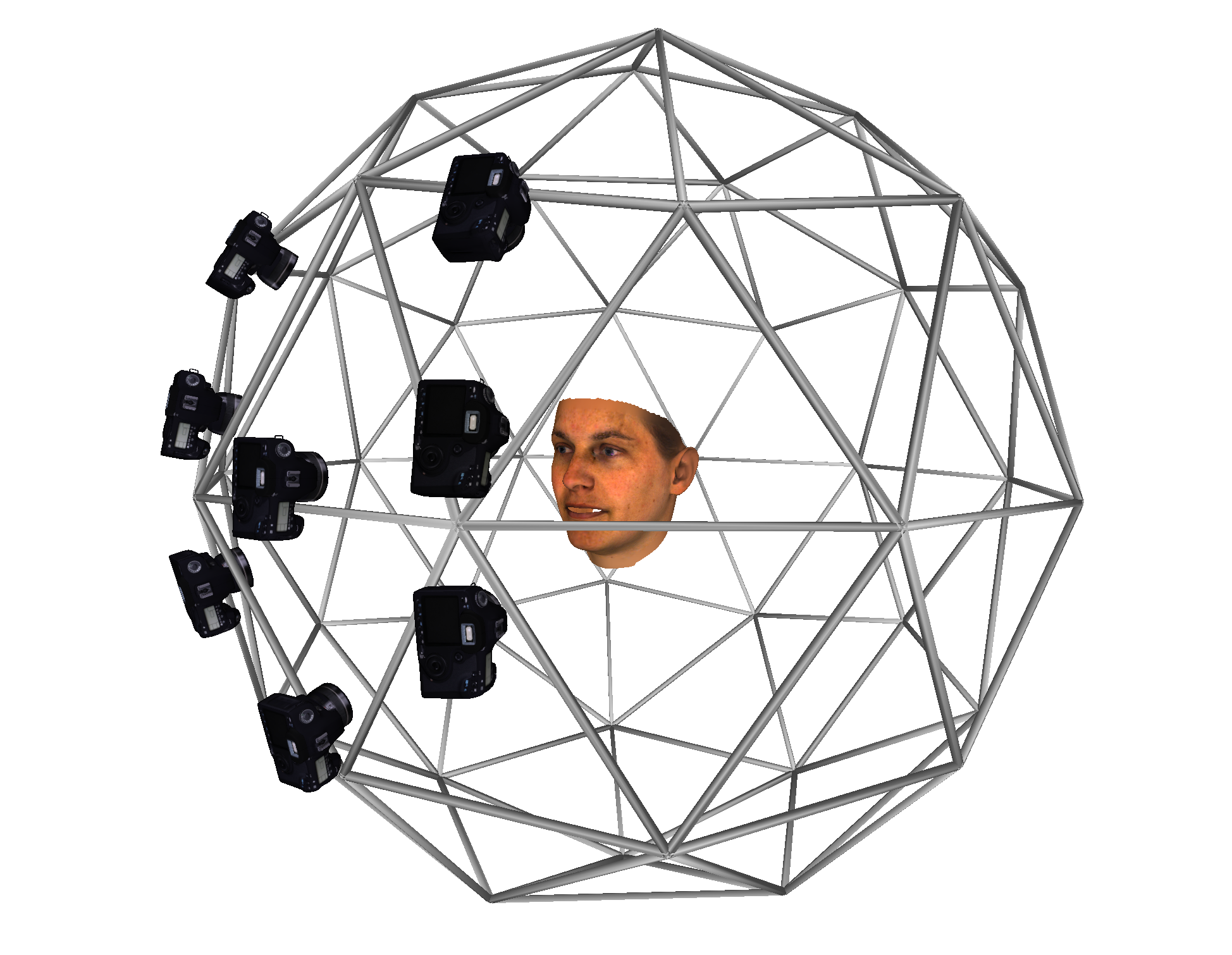}
    \includegraphics[trim=247px 72px 228px 31px,clip=true,width=5cm]{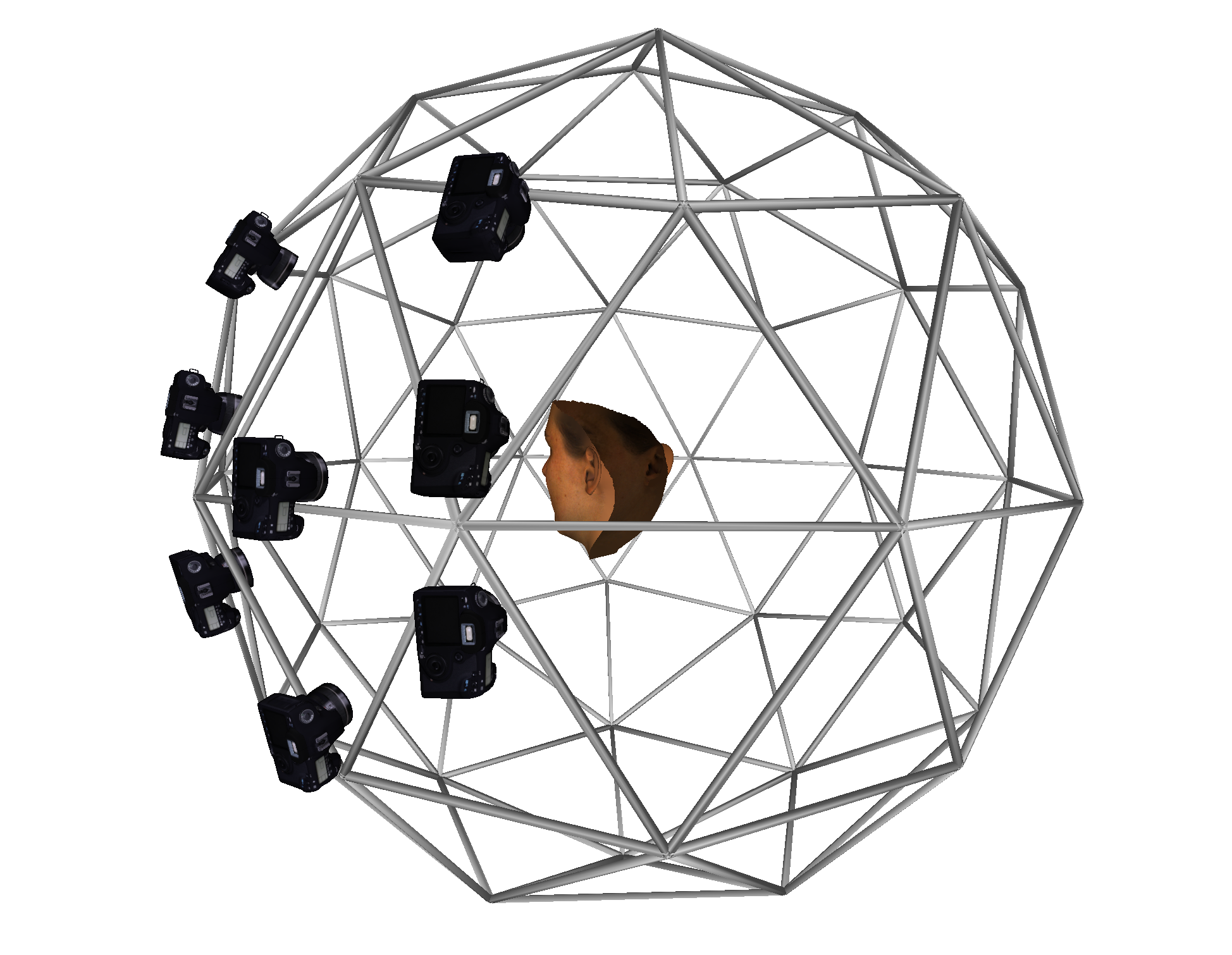}
    \caption{Positioning of cameras in our lightstage. We also show the three poses in which a face is captured, ultimately providing 24 views of the face.}
    \label{fig:setup}
\end{figure*}

In Figure \ref{fig:lightstage} we show a photo of our lightstage setup. The distribution of cameras is shown in Figure \ref{fig:setup} where we also show the three poses captured relative to the camera rig (approximately frontal and two profiles relative to the photometric camera). The camera positioned closest to the equator is our photometric camera for which the polarising filters are tuned. The other seven cameras simply provide additional viewpoints for multiview stereo.

\section{Demographics}

\begin{figure}
    \centering
    \includegraphics[trim=106px 305px 115px 320px,width=\columnwidth]{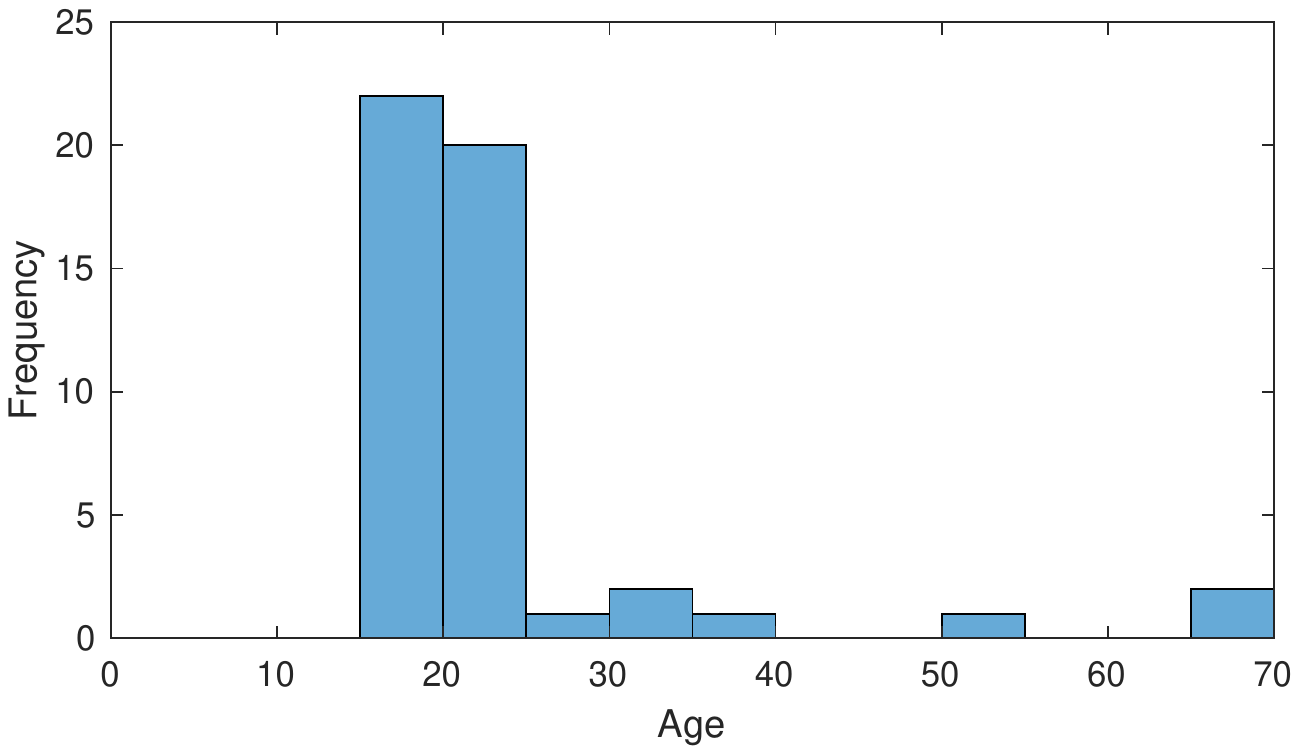}
    \caption{Age distribution for the 50 subjects we captured.}
    \label{fig:ages}
\end{figure}

\begin{figure}
    \centering
    \includegraphics[trim=106px 305px 115px 320px,width=\columnwidth]{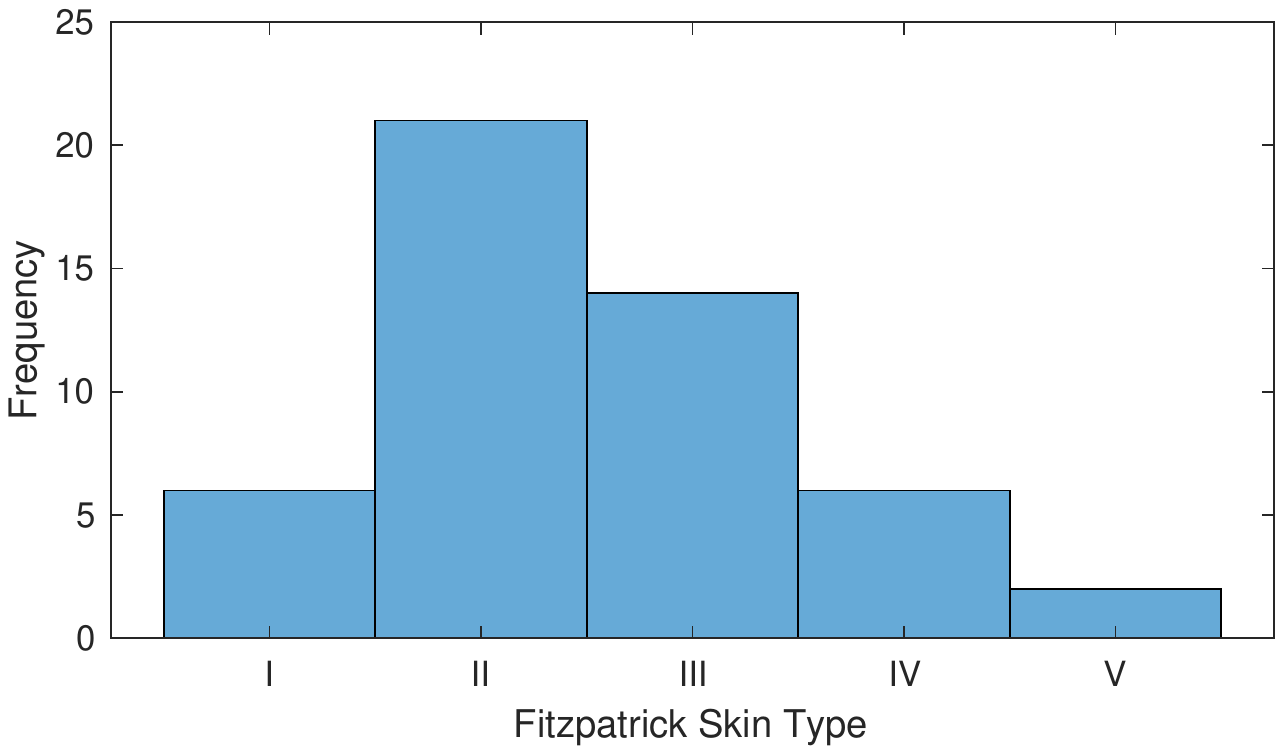}
    \caption{Skin type distribution for the 50 subjects we captured, according to the Fitzpatrick skin type classification.}
    \label{fig:skintypes}
\end{figure}

Our model is built from a total of 73 individuals (50 from our own captured data, 23 from the 3DRFE dataset \cite{stratou2011effect}). Of those, 17 are female (6 from 3DRFE). We do not have further demographic information on the 3DRFE participants but show the age distribution for the 50 participants we captured in Figure \ref{fig:ages} and skin type in Figure \ref{fig:skintypes}.

\section{Random model samples}

\begin{figure*}
    \centering
    \foreach \y in {1,2,3,4,5}
    {
        \foreach \x in {1,2,3,4,5,6,7,8,9,10}
        {    
            \includegraphics[width=0.09\textwidth]{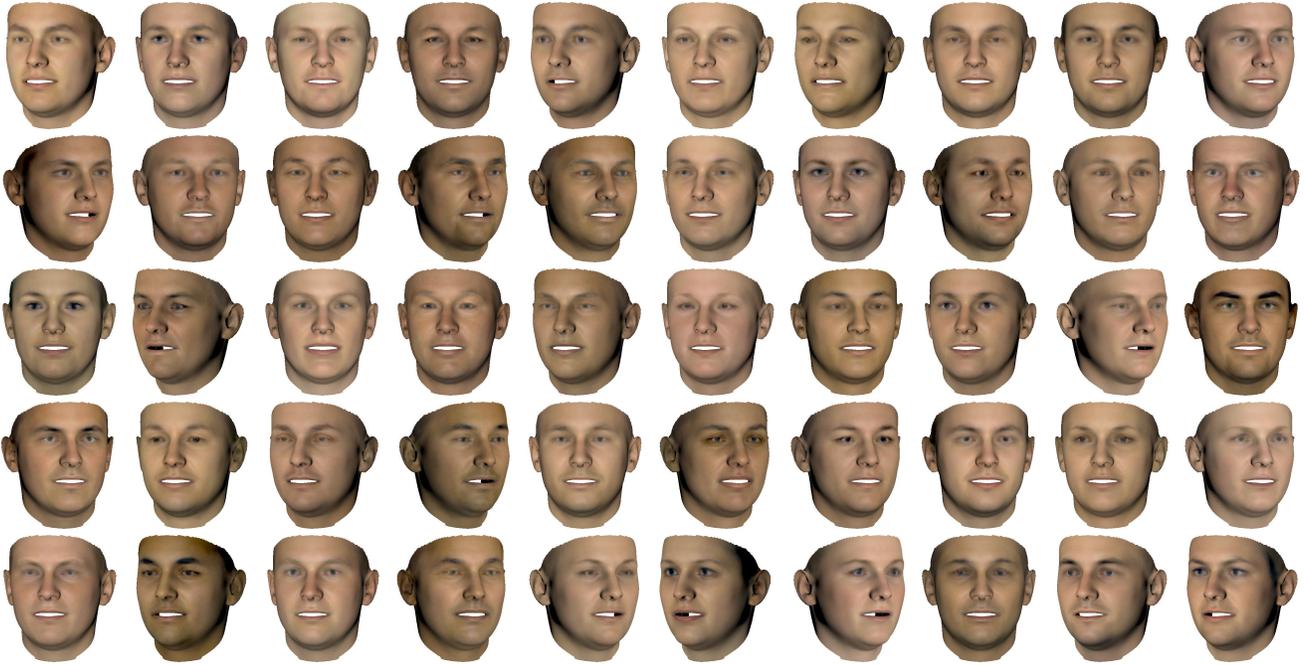}
        }
        \\
    }
    \caption{50 random samples drawn from our combined model and rendered in nonlinear sRGB space with white frontal point light source and Blinn-Phong reflectance with shininess set to 20. In all cases the shape is fixed to the BFM 2017 mean.}
    \label{fig:ours_random}
\end{figure*}

\begin{figure*}
    \centering
    \foreach \y in {1,2,3,4,5,6}
    {
        \foreach \x in {1,2,3,4,5,6,7,8,9,10}
        {    
            \includegraphics[width=0.09\textwidth]{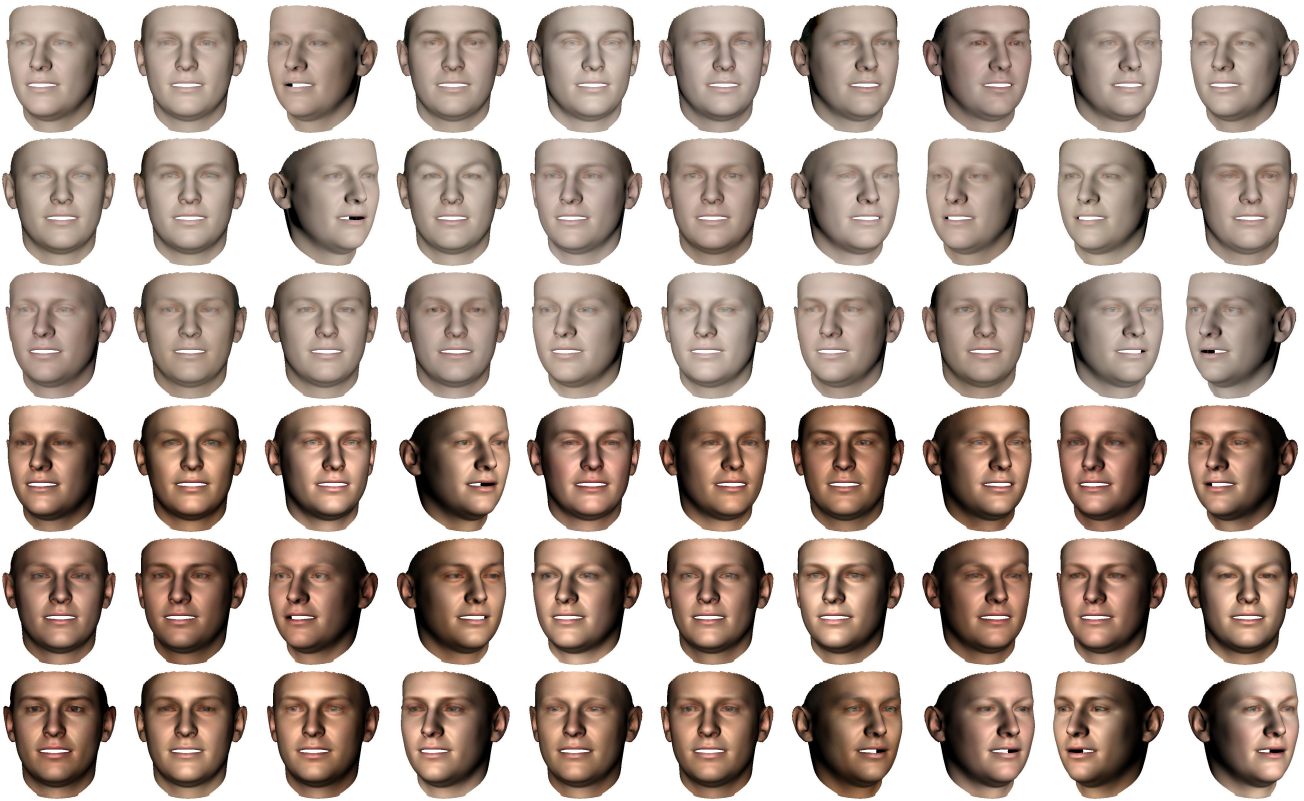}
        }
        \\
    }
    \caption{50 random samples drawn from the BFM 2017 \cite{gerig2018morphable}. We render with white frontal point light source and Blinn-Phong reflectance with shininess set to 20. In rows 1-3 we apply nonlinear gamma, in 4-6 we do not. We set the specular albedo to the average of the mean specular albedo map in our model. In all cases the shape is fixed to the BFM 2017 mean.}
    \label{fig:bfm_random}
\end{figure*}

\begin{figure*}
    \centering
    \foreach \y in {1,2,3,4,5,6}
    {
        \foreach \x in {1,2,3,4,5,6,7,8,9,10}
        {    
            \includegraphics[width=0.09\textwidth]{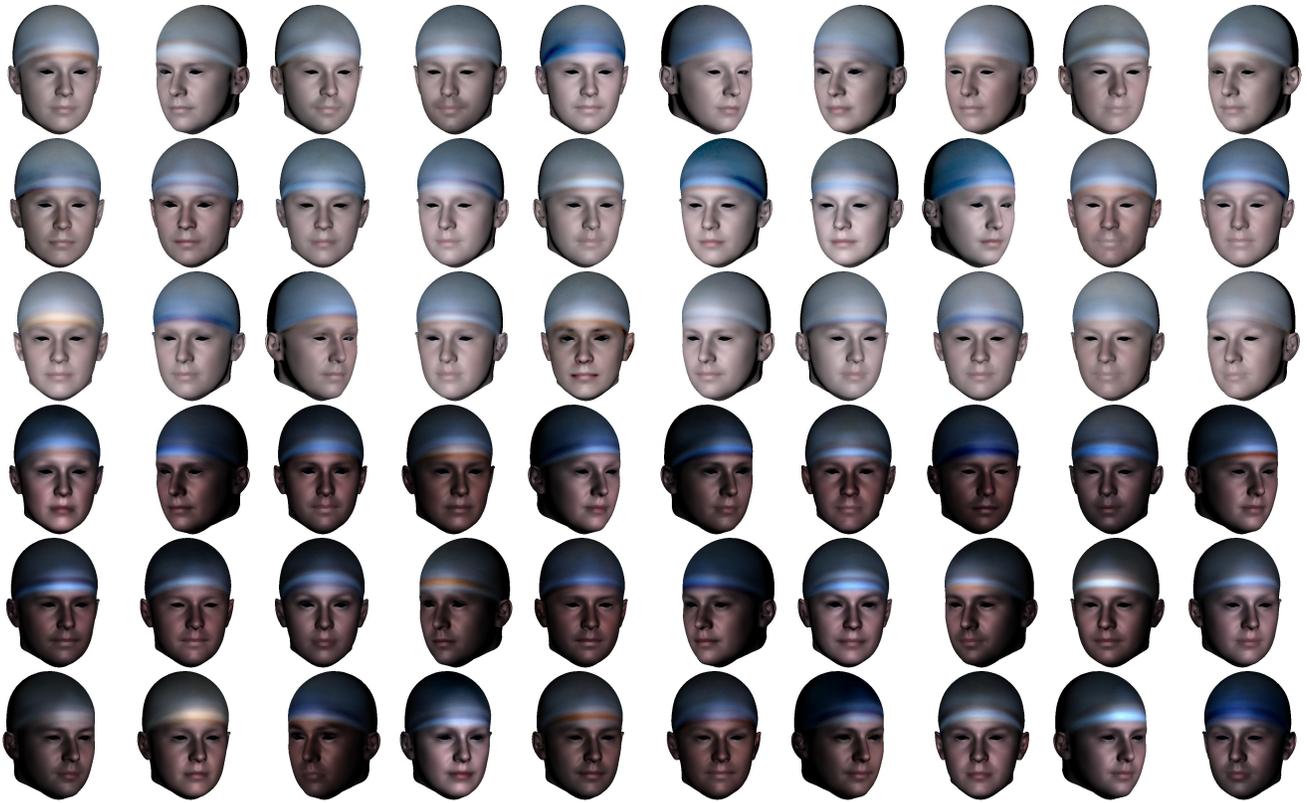}
        }
        \\
    }
    \caption{50 random samples drawn from the LYHM \cite{LYHM2017}. We render with white frontal point light source and Blinn-Phong reflectance with shininess set to 20. In rows 1-3 we apply nonlinear gamma, in 4-6 we do not. We set the specular albedo to the average of the mean specular albedo map in our model. In all cases the shape is fixed to the LYHM mean.}
    \label{fig:lyhm_random}
\end{figure*}

To provide another means to compare our model with current state-of-the-art, we draw random faces from each model and render them with a random rotation about the vertical axis drawn uniformly from $[-30^{\circ},30^{\circ}]$. In Figure \ref{fig:ours_random} we show 50 faces from our combined model. In Figure \ref{fig:bfm_random} we show 60 faces from the BFM 2017 \cite{gerig2018morphable}, half with nonlinear gamma applied, half without. The model was not intended to be used with gamma applied but we note that the shading unrealistically severe without. We provide the same visualisation for the LYHM \cite{LYHM2017} in Figure \ref{fig:lyhm_random}. Our model shows better diversity of skin colour and appearance while leading to model natural rendered appearance comparing to the other models either with or without gamma applied.

\section{Model visualisation in nonlinear space}

\begin{figure}[!t]
    \centering
\begingroup
\setlength{\tabcolsep}{1pt}
\renewcommand{\arraystretch}{0.5}
\begin{tabular}{m{0.3cm}m{1.2cm}c}
 \rotatebox{90}{Diffuse (nonlinear sRGB)} &
    \includegraphics[width=1.2cm]{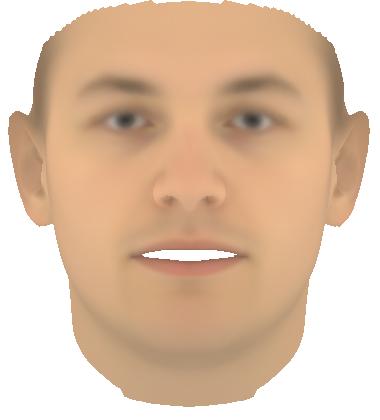} &
    \begin{tabular}{ccccc}
    \includegraphics[width=1.2cm]{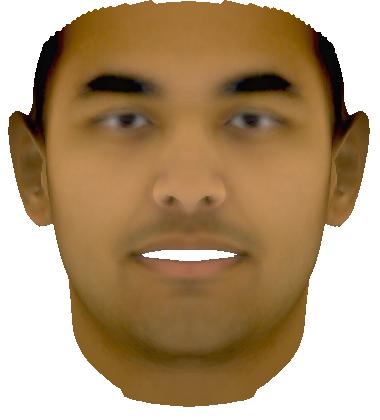} &
    \includegraphics[width=1.2cm]{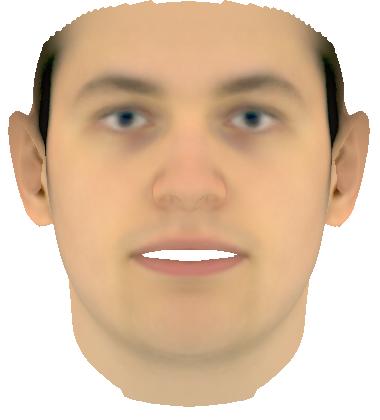} &
    \includegraphics[width=1.2cm]{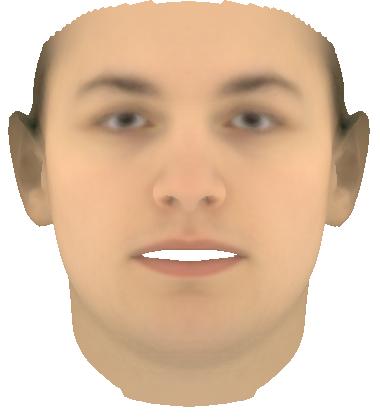} &
    \includegraphics[width=1.2cm]{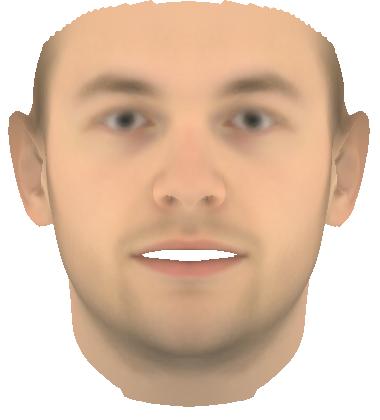} &
    \includegraphics[width=1.2cm]{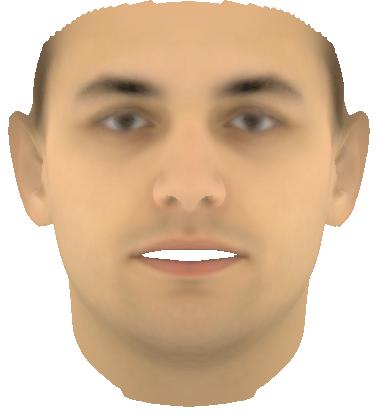} \\
    \includegraphics[width=1.2cm]{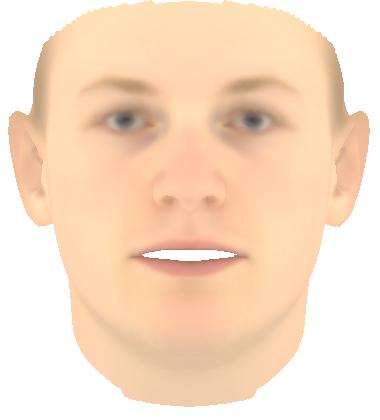} &
    \includegraphics[width=1.2cm]{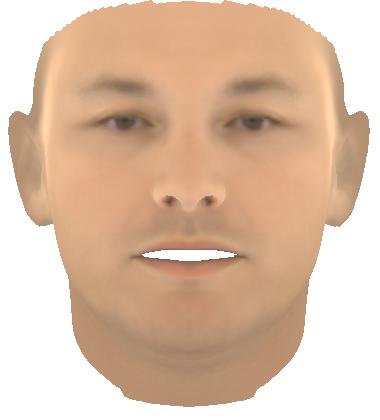} &
    \includegraphics[width=1.2cm]{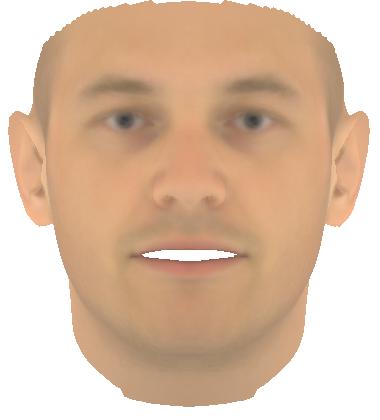} &
    \includegraphics[width=1.2cm]{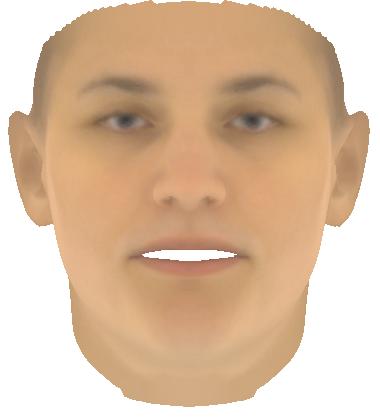} &
    \includegraphics[width=1.2cm]{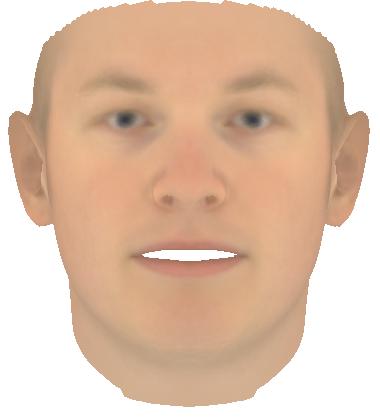} 
    \end{tabular} \vspace{0.1cm} \\
    \hline \\
    \rotatebox{90}{Specular (nonlinear sRGB)} &
    \includegraphics[width=1.2cm]{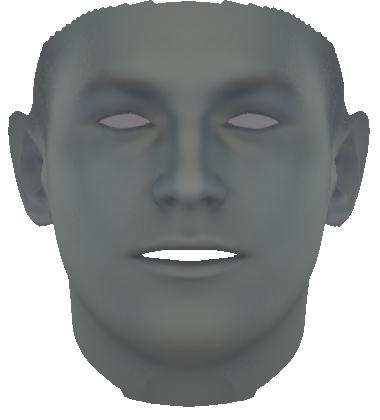} &
    \begin{tabular}{ccccc}
    \includegraphics[width=1.2cm]{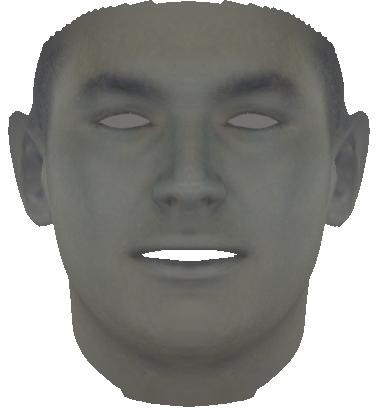} &
    \includegraphics[width=1.2cm]{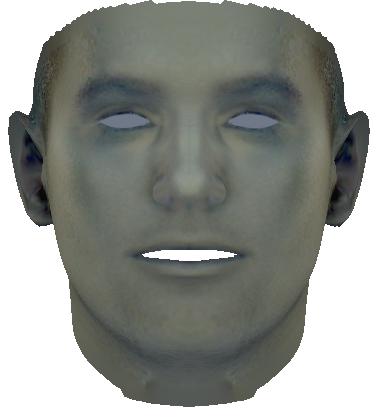} &
    \includegraphics[width=1.2cm]{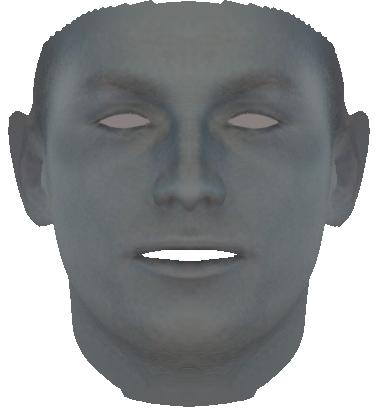} &
    \includegraphics[width=1.2cm]{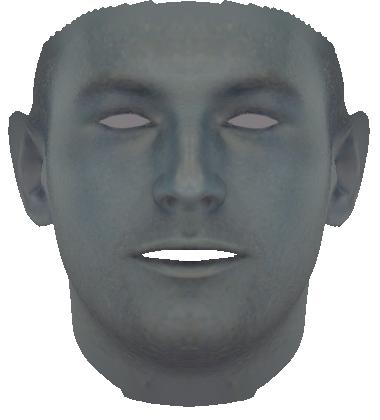} &
    \includegraphics[width=1.2cm]{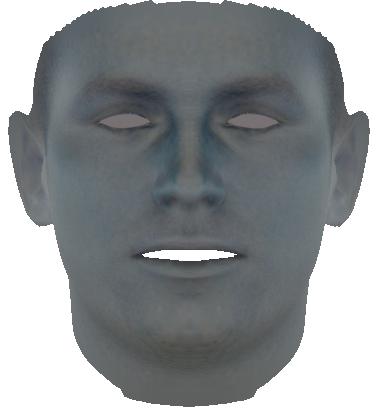} \\
    \includegraphics[width=1.2cm]{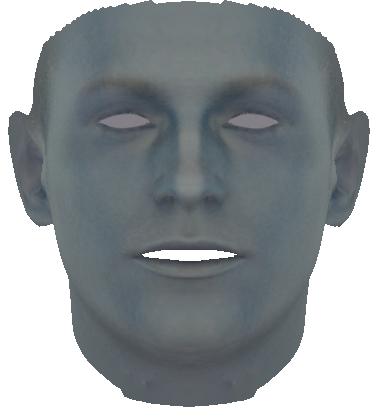} &
    \includegraphics[width=1.2cm]{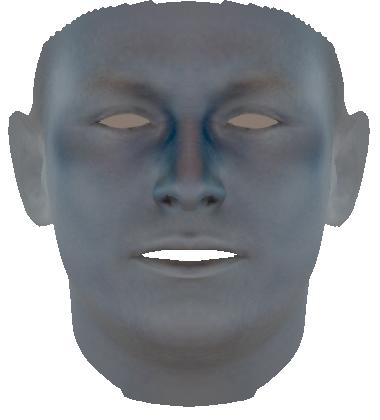} &
    \includegraphics[width=1.2cm]{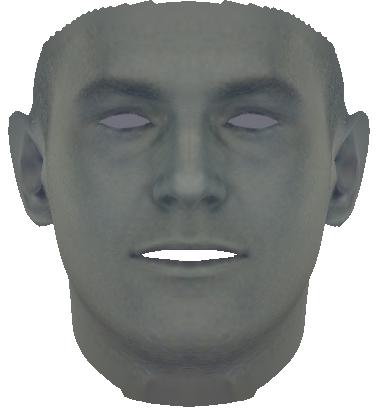} &
    \includegraphics[width=1.2cm]{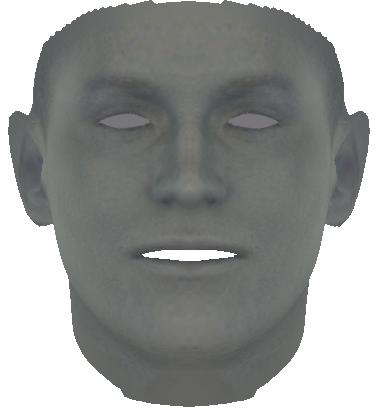} &
    \includegraphics[width=1.2cm]{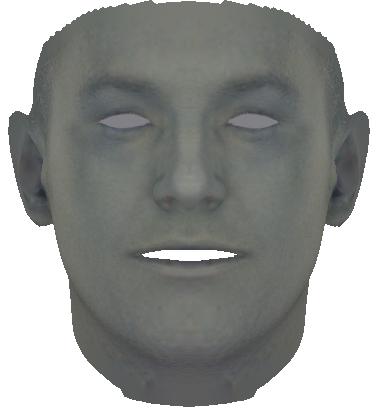} 
    \end{tabular}
\end{tabular}
\endgroup
    \caption{Visualisation of our diffuse and specular albedo models in nonlinear sRGB space.}
    \label{fig:modelgamma}
\end{figure}

A PCA model is a linear subspace model. Diffuse and specular albedo should be statistically modelled in a linear colour space with nonlinear gamma applied subsequently as part of the image formation model. However, we are not used to seeing either linear space face images nor pure albedo images and so the model visualisations look unnatural. To provide another visualisation of our model, in Figure \ref{fig:modelgamma} we show our diffuse and specular albedo models with nonlinear gamma applied, i.e.~in nonlinear sRGB space.

\section{Additional qualitative results}
\begin{figure*}[!t]
    \centering
    \includegraphics[width=\textwidth]{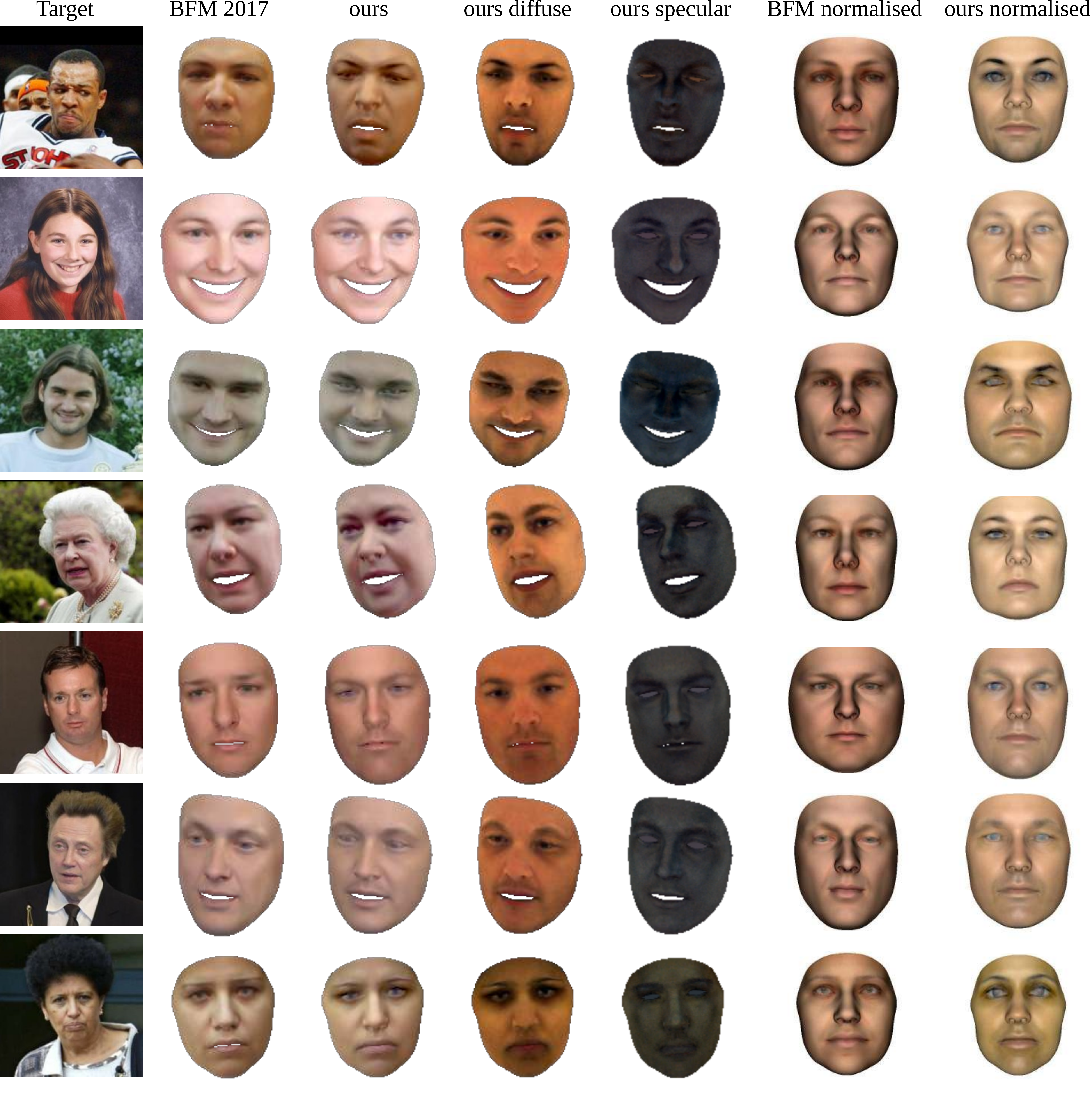}
    \caption{Qualitative model adaptation results on the LFW dataset \cite{LFWTech}. Our model leads to comparable results whilst explicitly disentangling albedo and estimating diffuse and specular albedo. We visualize both reconstructions under frontal pose and illumination in the normalised setting.}
    \label{fig:qualitative_supp}
\end{figure*}

We incorporate additional modalities to the figures in the paper to allow a better quantitative comparison of our inverse rendering results in Figure~\ref{fig:qualitative_supp} and~\ref{fig:benchmark}.

\begin{figure*}[!t]
    \centering
    \includegraphics[width=0.8\textwidth]{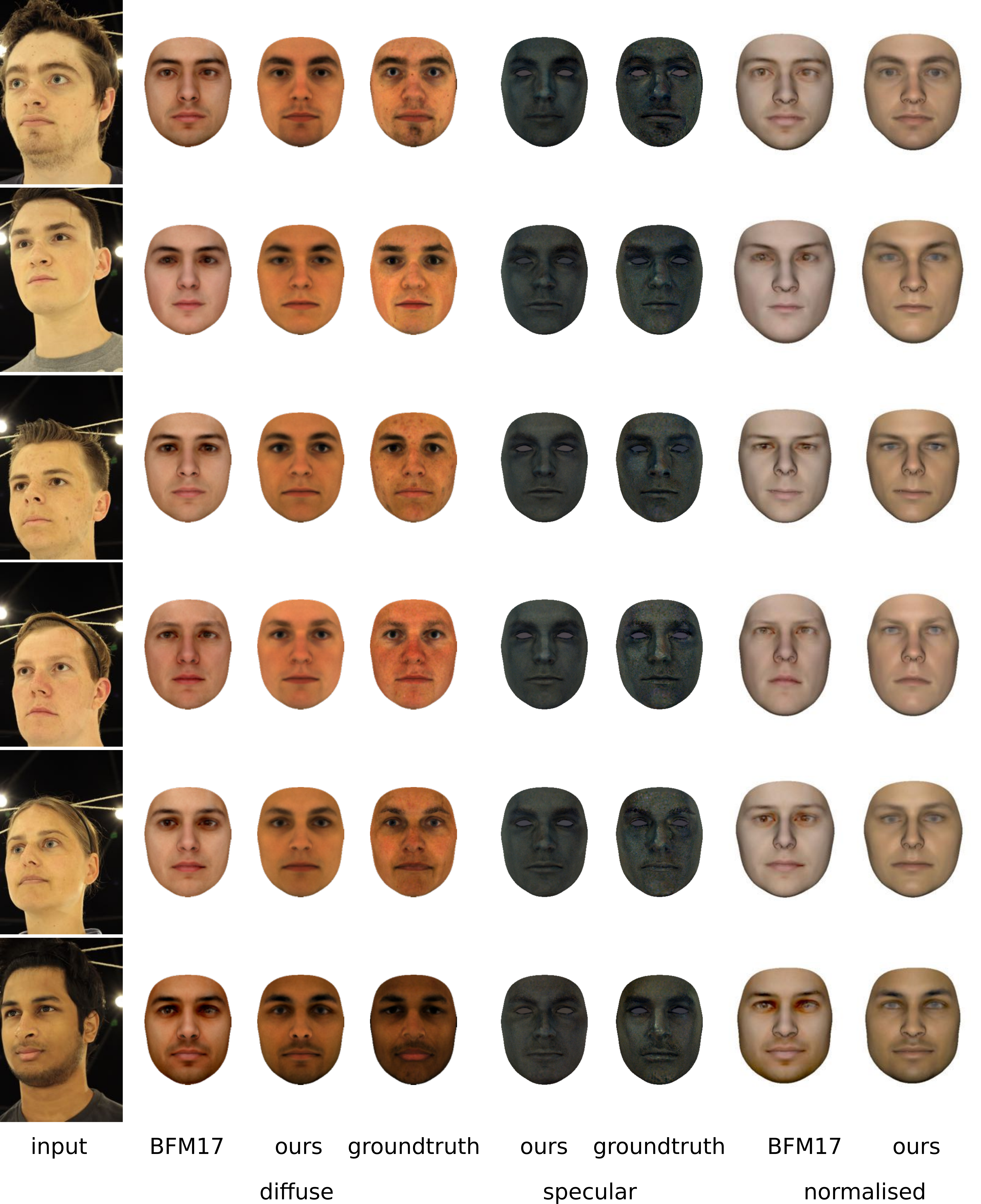}
    \caption{Albedo estimation results based on the exact same inverse rendering pipeline for the BFM 2017 and the proposed model. We present reconstructions of diffuse and specular albedo.}
    \label{fig:benchmark}
\end{figure*}

\section{Colour transformation}

\begin{figure}
    \centering
    \includegraphics[width=\columnwidth,trim=100px 304px 112px 316px,clip=true]{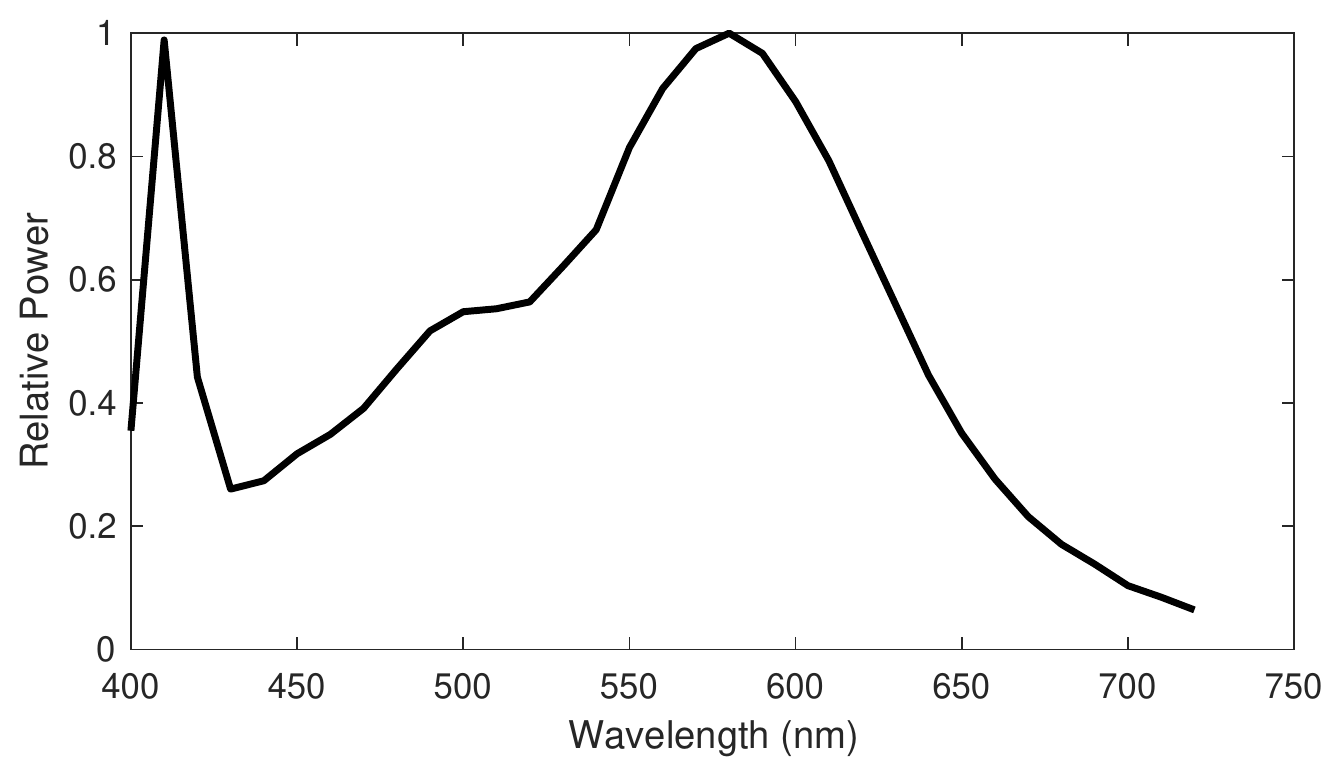}
    \caption{Spectral power distribution of LEDs used in our lightstage.}
    \label{fig:SPD}
\end{figure}

\begin{figure}
    \centering
    \includegraphics[width=\columnwidth,trim=100px 304px 112px 316px,clip=true]{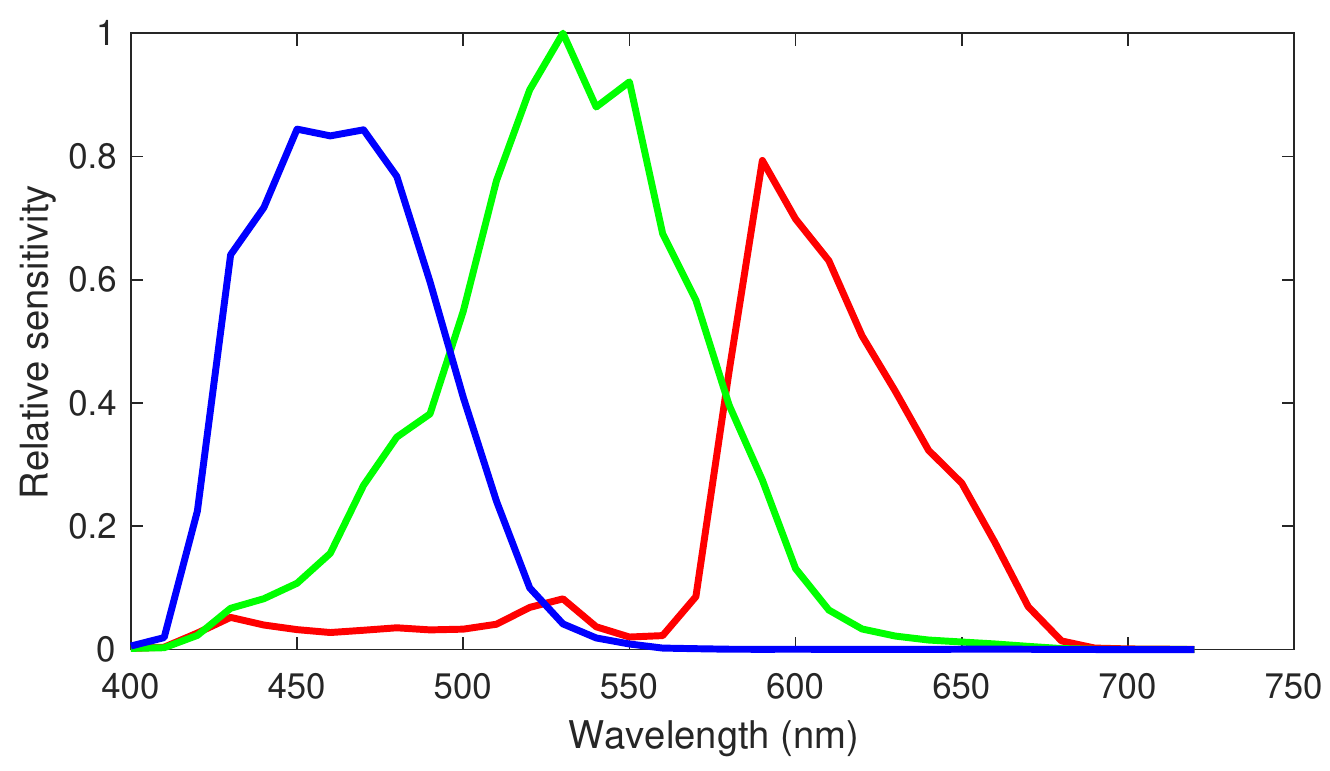}
    \caption{Spectral sensitivity of Nikon D200 used to capture photometric images.}
    \label{fig:specsens}
\end{figure}

In Figure \ref{fig:SPD} we show a plot of $\mathbf{e}$, the measured light source spectral power distribution for the LEDs used in our lightstage. Note that because our polarising filters are not spectrally neutral, we measure the SPD with the polarisers on the LED. In Figure \ref{fig:specsens} we show a plot of $\mathbf{C}_r$, $\mathbf{C}_g$ and $\mathbf{C}_b$, the camera spectral sensitivities for the Nikon D200 that we use to capture the photometric images. These were measured by Jiang~\etal~\cite{jiang2013space}. From these two measurements, we can compute an exact transformation from the RAW colour space captured by our camera to standardised sRGB space. Since most cameras provide images in this colour space, this means our model can be used to analyse images directly without further colour transformation. For applications in other colour spaces, we provide with our model the measured $\mathbf{e}$ and $\mathbf{C}$ as well as the derived components of the colour transformation matrix that we apply to the albedo maps.

\end{appendices}

\end{document}